\documentclass[10pt,twocolumn,letterpaper]{article}

\usepackage[pagenumbers]{cvpr} 

\definecolor{cvprblue}{rgb}{0.21,0.49,0.74}
\usepackage[pagebackref,breaklinks,colorlinks,allcolors=cvprblue]{hyperref}

\usepackage{booktabs}      
\usepackage{multirow}      
\usepackage{multicol}      
\usepackage{makecell}      
\usepackage{colortbl}      
\PassOptionsToPackage{table}{xcolor}
\usepackage{xcolor}        
\usepackage{threeparttable}
\usepackage{caption}
\usepackage{float}
\usepackage{listings}
\usepackage{tcolorbox}
\tcbuselibrary{skins, breakable}
\usepackage{stfloats}
\usepackage{fontawesome5}

\definecolor{codebg}{RGB}{250,250,250}
\definecolor{codeframe}{RGB}{220,220,220}
\definecolor{promptbg}{RGB}{252,252,252}
\definecolor{promptborder}{RGB}{180,180,180}

\lstdefinestyle{jsonstyle}{
    backgroundcolor=\color{codebg},
    basicstyle=\ttfamily\footnotesize,
    frame=single,
    rulecolor=\color{codeframe},
    frameround=tttt,
    tabsize=2,
    breaklines=true,
    showstringspaces=false
}

\definecolor{tableblue}{RGB}{200,220,255}

\newcommand{\bluecell}[2]{%
    \ifnum#1=100
      \cellcolor{tableblue!100!white}#2\%%
    \else
      \cellcolor{tableblue!40!white}#2\%%
    \fi
}

\newcommand{\bluecellnp}[2]{%
  \ifnum#1=100
    \cellcolor{tableblue!100!white}#2%
  \else
    \cellcolor{tableblue!40!white}#2%
  \fi
}

\newcommand{\legendlabel}[2]{%
  \begingroup
  \setlength{\fboxsep}{0.5pt}
  \colorbox{tableblue!#1!white}{\textbf{#2}}%
  \endgroup
}

\definecolor{lightpink}{RGB}{255,230,230}
\definecolor{lightpurple}{RGB}{240,230,255}
\definecolor{lightgreen}{RGB}{240,255,240}
\definecolor{lightred}{RGB}{255,200,185}
\definecolor{lightblue}{RGB}{255,245,200}
\definecolor{graytext}{gray}{0.3}
\definecolor{myred}{RGB}{220,0,0}
\definecolor{mygreen}{RGB}{0,150,0}
\definecolor{myblue}{RGB}{30,90,255}

\newtcolorbox{promptbox}[1][]{
    enhanced,
    breakable,
    colback=promptbg,
    colframe=promptborder,
    coltitle=white,
    colbacktitle=black,
    fonttitle=\bfseries,
    title=#1,
    boxrule=0.8pt,
    arc=2pt,
    left=8pt,
    right=8pt,
    top=8pt,
    bottom=8pt,
    boxsep=2pt,
    before skip=10pt,
    after skip=10pt,
}

\renewcommand{\arraystretch}{1.15}
\def\paperID{7858} 
\def\confName{CVPR}
\def\confYear{2026}

\title{UniPercept: Towards Unified Perceptual-Level Image Understanding across Aesthetics, Quality, Structure, and Texture}

\author{%
    {\large Shuo Cao$^{1,2,*,\diamondsuit}$, \hskip 0.8em Jiayang Li$^{3,*}$, \hskip 0.8em Xiaohui Li$^{2,4}$, \hskip 0.8em Yuandong Pu$^{2,4}$, \hskip 0.8em Kaiwen Zhu$^{2,4}$}\\%
    {\large Yuanting Gao$^{5}$, \hskip 0.8em Siqi Luo$^{2,4}$, \hskip 0.8em Yi Xin$^{2,6}$, \hskip 0.8em Qi Qin$^{2}$, \hskip 0.8em Yu Zhou$^{7}$}\\%
    {\large Xiangyu Chen$^{8}$, \hskip 0.8em Wenlong Zhang$^{2}$, \hskip 0.8em Bin Fu$^{2}$, \hskip 0.8em Yu Qiao$^{2}$, \hskip 0.8em Yihao Liu$^{2,\dagger}$}\\
    {\small $^1$ University of Science and Technology of China \quad $^2$ Shanghai AI Laboratory \quad $^3$ Peking University}\\%
    {\small $^4$ Shanghai Jiao Tong University \quad $^5$ Tsinghua University \quad $^6$ Nanjing University}
    {\small $^7$ Sun Yat-sen University \quad $^8$ Tele-AI}\\
    {\small
    \renewcommand{\arraystretch}{1.3}
    \begin{tabular}{l@{\hskip 1.2em}l}
        \raisebox{-0.35ex}{\includegraphics[height=1.2em]{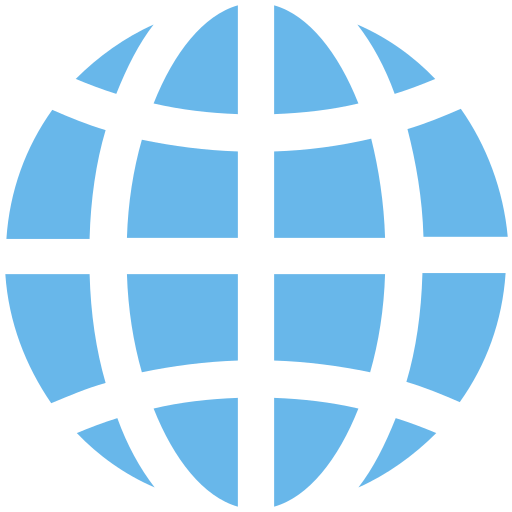}} \hspace{0.2em} \textbf{Website} & \url{https://thunderbolt215.github.io/Unipercept-project} \\
        \raisebox{-0.35ex}{\includegraphics[height=1.2em]{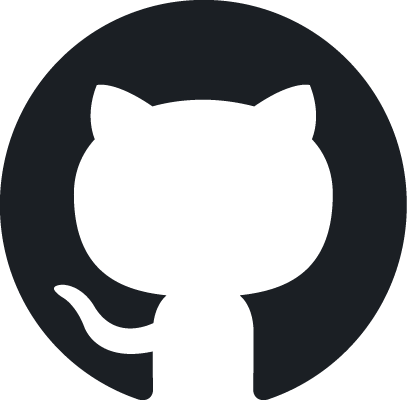}} \hspace{0.2em} \textbf{Code} & \url{https://github.com/thunderbolt215/UniPercept} \\
        \raisebox{-0.35ex}{\includegraphics[height=1.2em]{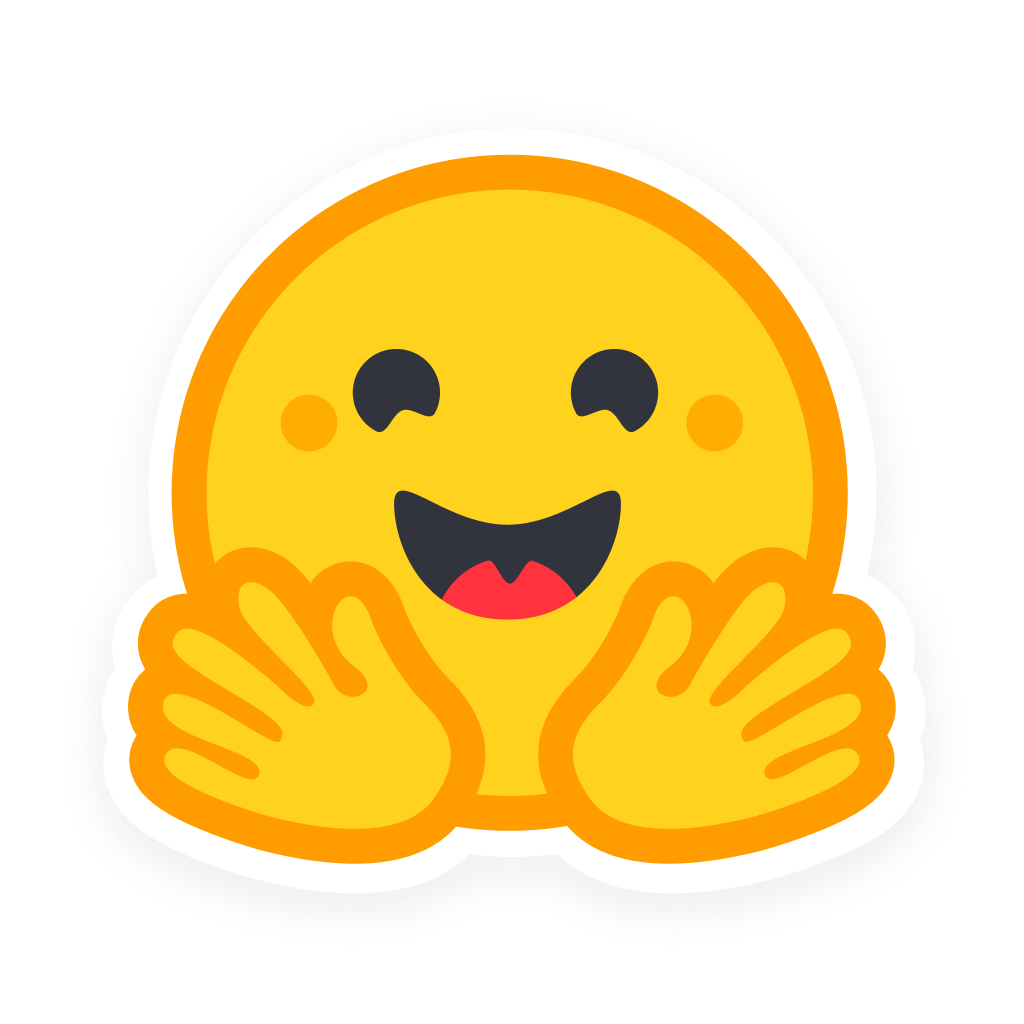}} \hspace{0.2em} \textbf{Benchmark \& Checkpoint} & \url{https://hf.co/collections/Thunderbolt215215/unipercept}
    \end{tabular}}\\[0.6cm]%
    \begin{minipage}{\textwidth}
        \centering
        \includegraphics[width=0.8\linewidth]{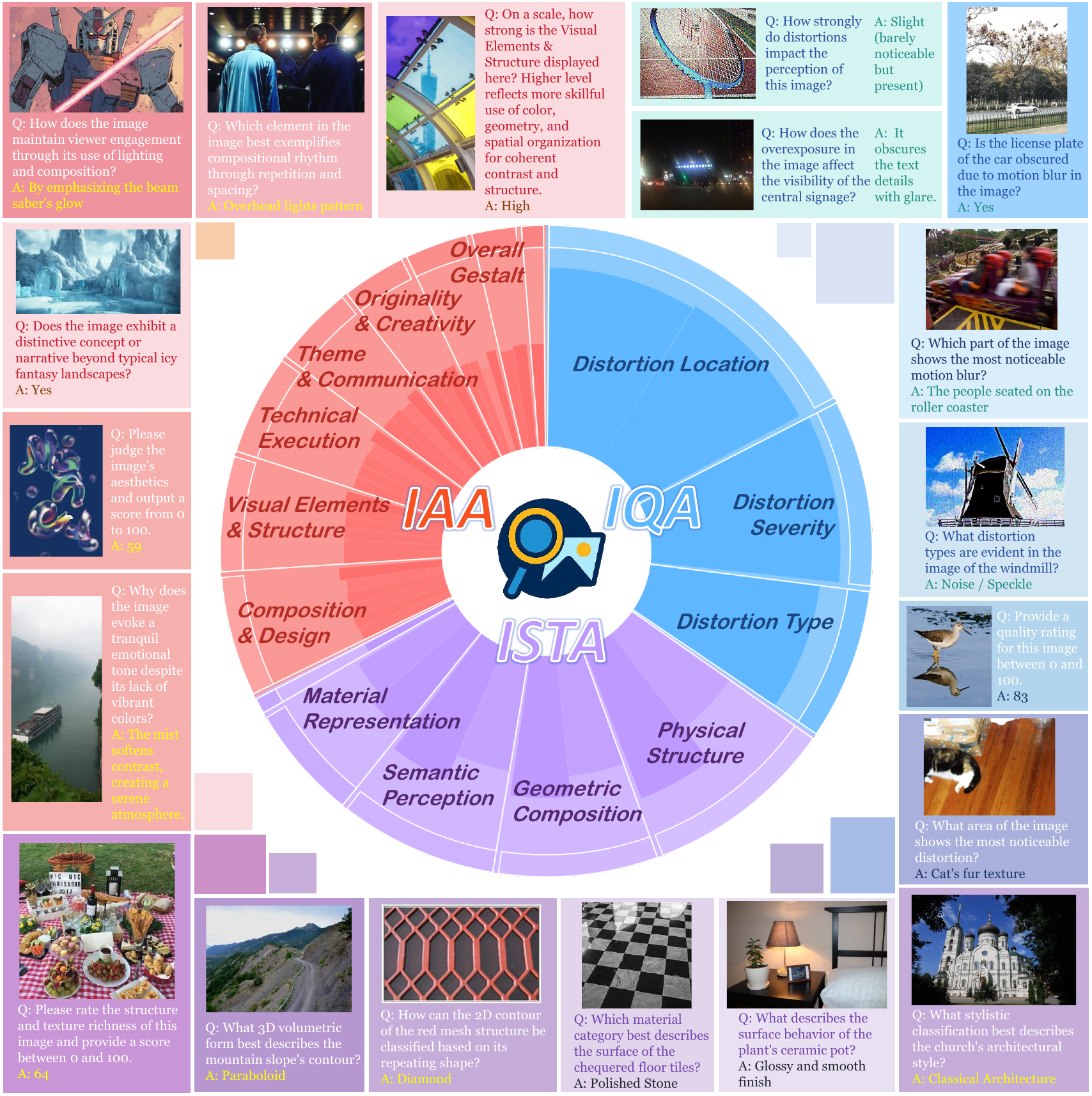}
        \vspace{-5pt}
        \captionof{figure}{\textbf{Overview of UniPercept-Bench.} 
        We propose a unified benchmark for perceptual-level image understanding, covering three perceptual domains: Image Aesthetics Assessment (\textbf{IAA}), Image Quality Assessment (\textbf{IQA}), and Image Structure and Texture Assessment (\textbf{ISTA}). 
        Each domain is organized hierarchically into \emph{Domain–Category–Criterion} levels, enabling fine-grained perceptual evaluation. 
        The benchmark contains both Visual Rating (\textbf{VR}) and Visual Question Answering (\textbf{VQA}) tasks, facilitating comprehensive assessment of models’ abilities to understand and reason about perceptual-level image understanding.}
        \label{fig:teaser}
    \end{minipage}
    \vspace{0.4cm} 
}

\begin{document}

\maketitle 

\begin{figure*}[b]
    \footnotesize
    \rule{5cm}{0.4pt} \\[-0.5pt] 
    $^*$Equal contribution. \quad 
    $^\diamondsuit$This work was done during his internship at Shanghai AI Laboratory. \\
    $^\dagger$Corresponding author.
\end{figure*}


\begin{abstract}
Multimodal large language models (MLLMs) have achieved remarkable progress in visual understanding tasks such as visual grounding, segmentation, and captioning. However, their ability to perceive \textbf{perceptual-level} image features remains limited. In this work, we present \textbf{UniPercept-Bench}, a unified framework for \textit{perceptual-level image understanding} across three key domains: \textbf{Aesthetics}, \textbf{Quality}, \textbf{Structure and Texture}. We establish a hierarchical definition system and construct large-scale datasets to evaluate perceptual-level image understanding. Based on this foundation, we develop a strong baseline \textbf{UniPercept} trained via Domain-Adaptive Pre-Training and Task-Aligned RL, enabling robust generalization across both \textbf{Visual Rating (VR)} and \textbf{Visual Question Answering (VQA)} tasks. UniPercept outperforms existing MLLMs on perceptual-level image understanding and can serve as a \textbf{plug-and-play reward model} for text-to-image generation. This work defines Perceptual-Level Image Understanding in the era of MLLMs and, through the introduction of a comprehensive benchmark together with a strong baseline, provides a solid foundation for advancing perceptual-level multimodal image understanding.
\end{abstract}

\section{Introduction}
\label{sec:intro}

Recent years have witnessed the rapid advancement of multimodal large language models (MLLMs), which now achieve impressive performance across a variety of vision–language tasks including segmentation, visual grounding, image captioning, and visual reasoning~\cite{gpt4, gemini, claude}. These advancements are largely driven by their strong capability to learn and align semantic-level representations, allowing models to identify objects and scenes, capture their relationships and perform visual reasoning~\cite{internvl3,qwen2.5,internvl3-5}.

\begin{figure}[htbp]
    \centering
    \includegraphics[width=0.45\textwidth]{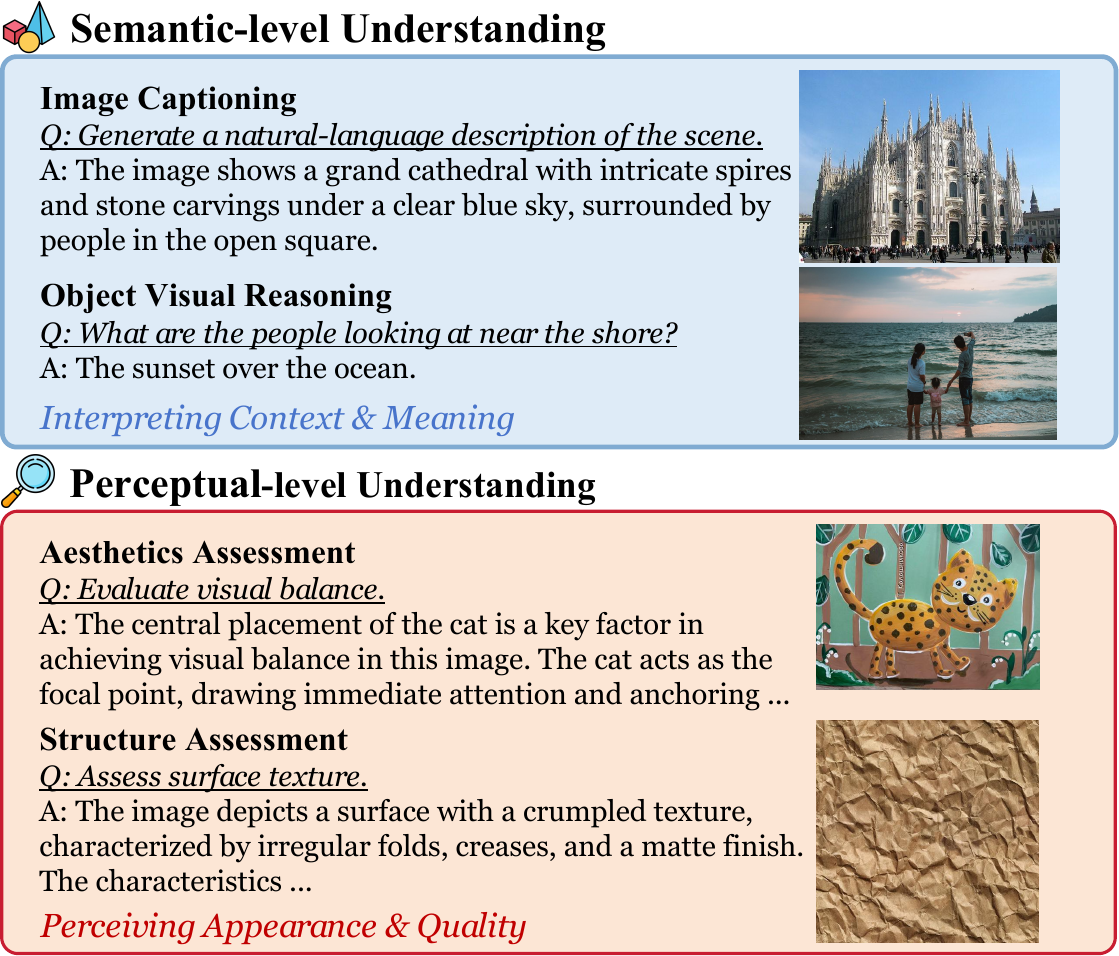}
    \vspace{-5pt}
    \caption{\textbf{Semantic-level} vs. \textbf{Perceptual-level} understanding.}
    \label{fig:task_compare}
\end{figure}

However, despite extensive progress on semantic understanding, the perceptual-level comprehension of images—how humans perceive aesthetics, quality, structure, and texture—remains substantially underexplored. As illustrated in Fig.~\ref{fig:task_compare}, \textbf{semantic-level} tasks focus on high-level interpretation of visual entities (e.g., object attributes or contextual reasoning), whereas \textbf{perceptual-level} tasks require assessing fine-grained, low-level visual appearance, such as aesthetic harmony, degradation severity, structural regularity, or surface texture. These perceptual attributes are inherently subtle, often subjective, and closely tied to human visual experience, making them fundamentally different from typical semantic-level tasks.


Human visual perception involves much more than object recognition: it includes nuanced judgments about \textit{how} an image looks and feels. Such perceptual cues play a crucial role in many downstream applications (e.g., content creation, image enhancement, and generative model alignment). Nevertheless, current MLLMs often struggle with these aspects, producing unstable or inconsistent predictions when evaluating aesthetic quality, perceptual degradation, or structural coherence. This gap highlights the need for a unified framework that can \textit{explicitly define, evaluate, and improve} perceptual-level understanding in MLLMs, as perceptual attributes remain far less standardized and less explored than semantic ones. Addressing this missing layer is essential for building models that align more closely with human judgments, thereby achieving higher visual quality.

To bridge this gap, we propose \textbf{UniPercept}, the first unified framework for \textbf{perceptual-level image understanding} across three core domains: Image Aesthetics Assessment (\textbf{IAA}), Image Quality Assessment (\textbf{IQA}), and Image Structure and Texture Assessment (\textbf{ISTA}). Our contributions are summarized as follows:

\begin{figure*}[t]
    \centering
    \includegraphics[width=\textwidth]{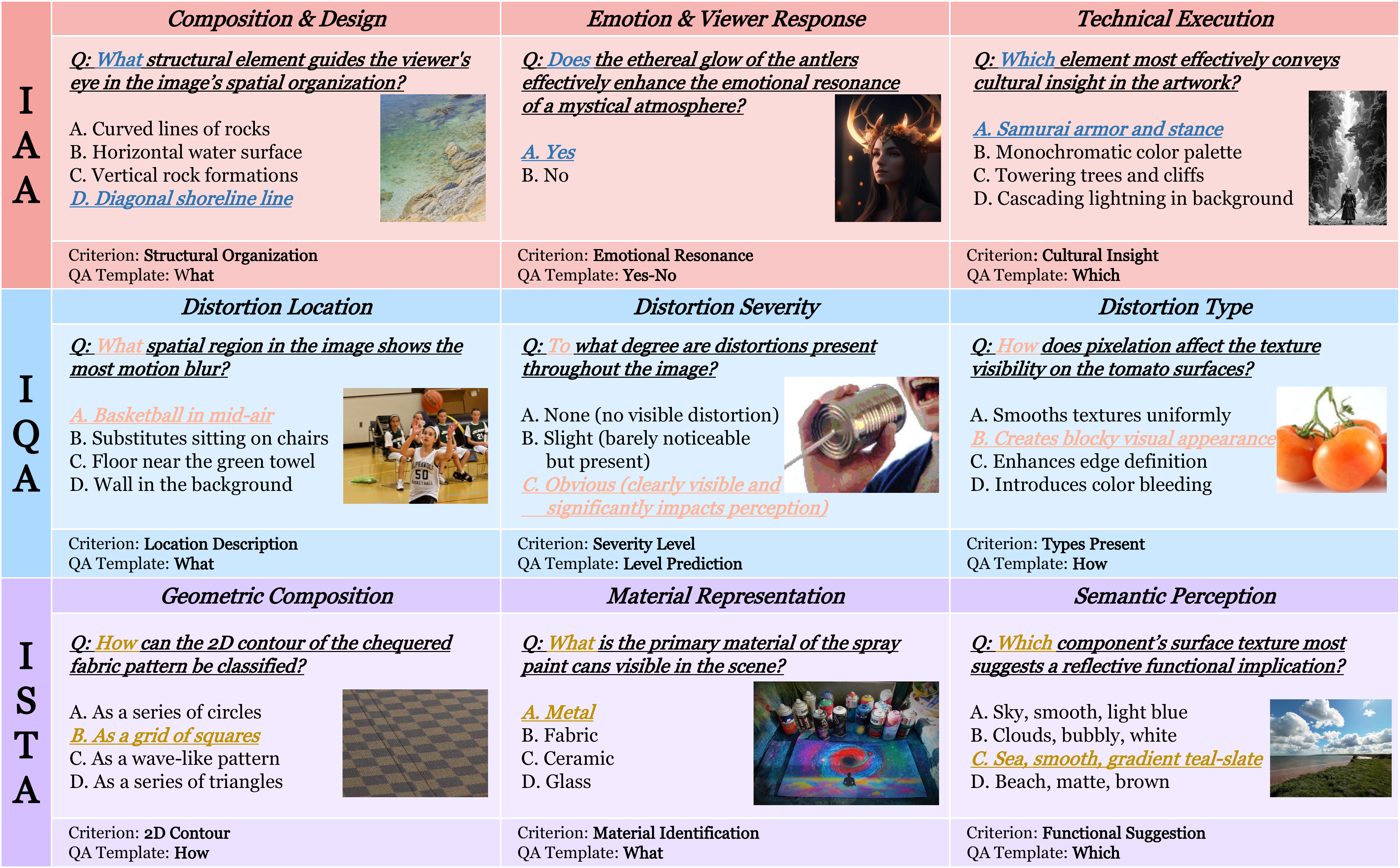}
    \vspace{-5pt}
    \caption{\textbf{Representative QA examples in UniPercept-Bench.}
Questions follow a three-level hierarchy of \textit{Domain–Category–Criterion}, defining perceptual scope, specific visual aspects, and fine-grained criteria for constructing diverse, perception-oriented VQA tasks.}
    \label{fig:qa_example}
\end{figure*}

\begin{itemize}
    \item \textbf{UniPercept-Bench.}  
    We establish a comprehensive hierarchical taxonomy of perceptual attributes, consisting of three progressive layers: \textit{Domain–Category–Criterion}. Building upon this taxonomy, we construct UniPercept-Bench, a systematically designed benchmark for evaluating perceptual-level image understanding in MLLMs. The benchmark covers fine-grained perceptual attributes and supports both Visual Rating (\textbf{VR}) and Visual Question Answering (\
    \textbf{VQA}) tasks for unified perceptual-level image understanding.

    \item \textbf{UniPercept.}  
    We develop \textbf{UniPercept} as a strong baseline MLLM through large-scale Domain-Adaptive Pre-Training and task-aligned reinforcement learning. Without relying on additional human feedback, the model learns to reliably assess perceptual attributes across diverse visual domains. UniPercept demonstrates strong generalization across both \textbf{VR} and \textbf{VQA} tasks, and achieves consistent gains in all three perceptual domains (\textbf{IAA}, \textbf{IQA}, and \textbf{ISTA}), substantially outperforming state-of-the-art generalized and specialized MLLMs.

    \item \textbf{Applications of UniPercept.}
UniPercept serves as a strong plug-and-play \textbf{reward model} for post-training T2I models~\cite{liu2025flowgrpotrainingflowmatching}, enabling direct optimization of perceptual-level signals such as aesthetic quality, structural richness, and scene diversity. This integration yields clear and controllable improvements in the perceptual quality of generated images.
Beyond reward optimization, UniPercept also functions as a unified perceptual metric for evaluating images and as a tool for characterizing perceptual distributions in large-scale datasets.
    
\end{itemize}

\section{Related Works}
\label{sec:related_work}

\subsection{MLLM Benchmark}
\label{sec:mllm_benchmark}

With the rapid development of MLLMs, evaluating their performance has gone far beyond simple semantic understanding tasks such as image recognition or segmentation. In recent years, researchers have begun to assess whether a model can achieve a deeper level of understanding and reasoning about visual content.

For instance, MMMU~\cite{mmmu} focuses on university-level exam questions across diverse disciplines, while MMMU-Pro~\cite{mmmu-pro} extends this idea with more complex, cross-domain reasoning challenges. MEGA-Bench~\cite{mega-bench} emphasizes large-scale multimodal comprehension and knowledge integration, and MMStar~\cite{mm-star} targets general reasoning and factual understanding in visual contexts. MMBench~\cite{mmbench} evaluates comprehensive perception and reasoning across everyday images, MathVista~\cite{mathvista} centers on mathematical and geometric reasoning within visual scenes, and OCRBench~\cite{ocrbench} specifically tests a model’s capability to recognize and interpret text embedded in images.


However, these benchmarks rely on converting visual content into text representations before reasoning, emphasizing language-based inference over genuine visual understanding. In contrast, UniPercept-Bench directly evaluates perceptual-level visual properties—such as technical execution, distortion location, and material depiction—bridging the gap between perceptual and semantic understanding.

\begin{figure*}[t]
    \centering
    \includegraphics[width=\textwidth]{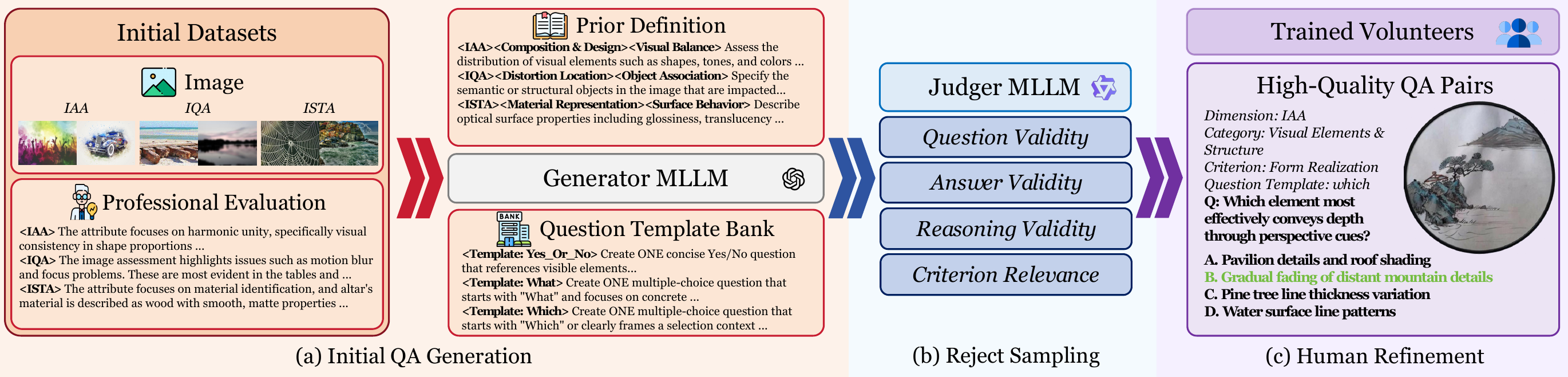}
    \vspace{-5pt}
    \caption{\textbf{Constuction pipeline of UniPercept-Bench.} A three-stage process initial QA generation, reject sampling, and human refinement to produce high-quality perceptual-level QA pairs across aesthetics, quality, structure, and texture. }
    \label{fig:benchmark_pipeline}
\end{figure*}

\begin{figure}[t]
    \centering
    \includegraphics[width=0.45\textwidth]{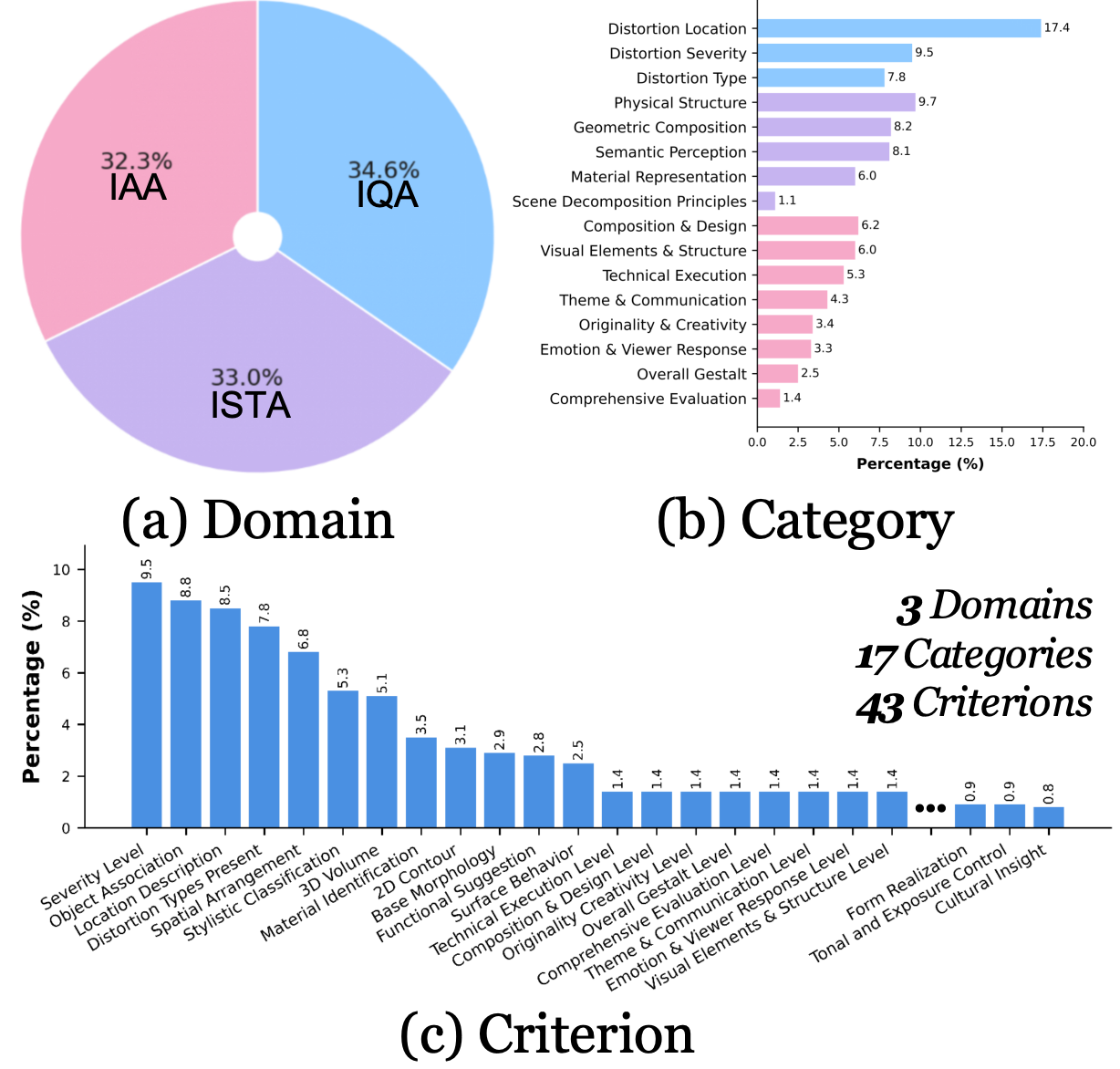}
    \vspace{-5pt}
    \caption{\textbf{Distribution of UniPercept-Bench} across (a) Domain, (b) Category, and (c) Criterion. Zoom in for best view.}
    \label{fig:benchmark_distribution}
\end{figure}


\subsection{Image Assessment}
\label{sec:image_assessment}


For perceptual-level image assessment, prior research has primarily focused on two major areas: 
Image Aesthetics Assessment (IAA) and 
Image Quality Assessment (IQA). 
Extensive benchmarks and methods have been developed for these two categories—for example, 
Q-Align~\cite{qalign}, UNIAA~\cite{Uniaa}, and ArtiMuse~\cite{artimuse} for IAA, and 
MUSIQ~\cite{Musiq}, DepictQA~\cite{depictqa_v1,depictqa_v2}, DeQA~\cite{deqa_score}, and Q-Insight~\cite{Q-insight} for IQA. In contrast, another crucial perceptual dimension—Image Structure and Texture Assessment (\textbf{ISTA})—has received far less systematic attention. 
Although a few prior works~\cite{dtd,fmd} touch upon aspects of structural or textural perception, they do not provide a unified or comprehensive definition of ISTA, leaving this important component of perceptual-level understanding still insufficiently explored.

In addition, most existing datasets focus on a single aspect such as numerical scoring or question answering, without providing a comprehensive and multi-dimensional evaluation framework. As multimodal large models continue to improve, many existing models have already achieved very high accuracy on prior benchmarks~\cite{Aesbench,Q-bench,Q-bench++}, reducing the ability of these benchmarks to effectively distinguish stronger models. In contrast, our UniPercept-Bench addresses these limitations by covering multiple perceptual dimensions, offering diverse and detailed evaluation data, and presenting a more comprehensive challenge for MLLMs.

\section{UniPercept-Bench}
\label{sec:benchmark}

\subsection{Definition}
\label{sec:definition}

As shown in Fig.~\ref{fig:teaser}, UniPercept-Bench targets \textbf{perceptual-level} image understanding across Image Aesthetics Assessment (IAA), Image Quality Assessment (IQA), and Image Structure and Texture Assessment (ISTA). \textbf{IAA} focuses on the \textit{perceived aesthetic attributes} of an image, such as composition, style, emotion, and overall visual appeal. \textbf{IQA} targets the \textit{perceived fidelity and degradation factors}, including noise, blur, compression artifacts, and overall distortion levels. \textbf{ISTA} evaluates the \textit{structural and textural characteristics} of a scene, emphasizing geometry, material properties, and local detail richness. Although all three domains assess images, the tasks focus on fundamentally \textbf{different aspects}. For example, a high-quality image may not possess strong aesthetic value, while an aesthetically pleasing image may contain only simple or sparse textures.

While IAA and IQA have been widely studied in prior works such as Q-Align~\cite{qalign}, the DepictQA series~\cite{depictqa_v1,depictqa_v2,deqa_score}, and ArtiMuse~\cite{artimuse}, \textbf{ISTA} remains largely unexplored, with little prior effort to systematically define or evaluate it. Inspired by recent advances in image generation and low-level vision~\cite{flux,wu2025qwenimagetechnicalreport,supir}, we reorganize and unify the existing definition systems for IAA and IQA, and propose the \textbf{first} systematic, operational definition of ISTA. Together, these three domains form a coherent perceptual-level framework that enables comprehensive and fine-grained assessment of IAA, IQA ans ISTA cues aligned with human perception. More details are provided in the Appendix.



\begin{figure*}[htbp]
    \centering
    \includegraphics[width=0.9\textwidth]{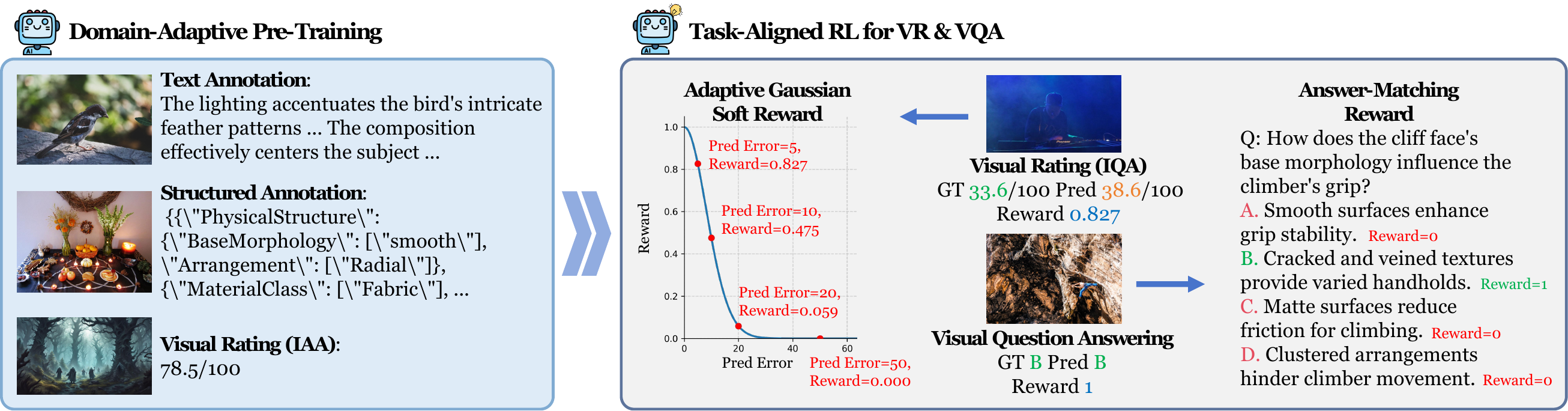}
    \vspace{-5pt}
    \caption{\textbf{Training pipeline of UniPercept.} 
A two-stage framework combining domain-adaptive pre-training for perceptual understanding and task-aligned RL to jointly optimize Visual Rating and Visual Question Answering.}
    \label{fig:training_pipeline}
\end{figure*}

\subsection{Benchmark Construction}
\label{sec:benchmark_construction}



We organize UniPercept-Bench using a three-tier taxonomy of \textit{Domain–Category–Criterion}.
For IAA and IQA, we consolidate expert-agreed definitions from prior literature.
For ISTA, we conduct structured interviews with domain practitioners to derive precise, consensus-driven perceptual criteria. Building on this taxonomy, UniPercept-Bench introduces two complementary task forms:


\noindent\textbf{Visual Rating (VR).} 
Models output a continuous score representing perceptual alignment across IAA (aesthetics quality) and IQA (image quality), serving as a quantitative measure of perceptual understanding. 
We additionally define the \textbf{ISTA Rating} task to evaluate \textit{structure and texture richness}, where a higher score indicates a scene with more complex geometry and richer textures. 
Following the definition framework of UniPercept-Bench, we design a scoring method that computes the ISTA score by aggregating and weighting the counts of structured attributes extracted from each image, as detailed in the Appendix.

\noindent\textbf{Visual Question Answering (VQA).} Question sets are constructed at the \textit{Domain–Category–Criterion} levels to assess perceptual-level understanding and reasoning. Together, VR and VQA form a unified evaluation protocol that jointly measures \textbf{quantitative judgment} and \textbf{explanatory consistency}, advancing comprehensive perceptual-level understanding, as illustrated in Fig.~\ref{fig:qa_example}.



\noindent\textbf{Initial QA Generation.}
As illustrated in Fig.~\ref{fig:benchmark_pipeline}, We begin by collecting images from three perceptual dimensions to ensure diversity across aesthetic quality, distortion severity, and structural complexity. For IAA and IQA, we adopt expert-annotated datasets~\cite{artimuse,depictqa_v2,q-ground} as professional evaluations. 
For ISTA, we introduce structured annotations guided by UniPercept-Bench’s taxonomy, describing scene composition, geometric layout, and material representation.
These annotations are paired with predefined \textit{Domain–Category–Criterion} definitions and matched templates from a curated \textit{Question Template Bank}. The \textit{Generator MLLM} (GPT-4o~\cite{gpt4}) combines the image, annotation, and template to produce candidate QA pairs along with brief reasoning rationales.

\noindent\textbf{Reject Sampling.} A heterogeneous \textit{Judger MLLM} (Qwen-2.5-VL-78B-Instruct~\cite{qwen2.5}) evaluates each QA pair on \textbf{Question Validity}, \textbf{Answer Validity}, \textbf{Reasoning Validity}, and \textbf{Criterion Relevance}using a five-point scale. Only samples rated \textit{good} or above on all aspects are retained, removing about 40\% of candidates.


\noindent\textbf{Human Refinement.}
Finally, trained volunteers with expertise in Image Assessment conduct manual validation to ensure alignment with human reasoning and perception. 
Borderline cases are revised for clarity and consistency, while invalid samples are removed. 
The resulting dataset consists of high-quality QA pairs that are both perceptually grounded and semantically coherent.


\section{UniPercept}
\label{sec:baseline}

\subsection{Domain-Adaptive Pre-Training}
\label{sec:pretrain}

We aim to train a model with perceptual-level understanding ability across Aesthetics, Quality, Structure, and Texture. To endow the model with a foundational capability for perceptual-level image understanding, we first conduct a \textbf{Domain-Adaptive Pre-Training} stage. Specifically, we select and process large-scale existing datasets, resulting in two types of pre-training data as follows:


\noindent\textbf{Text-based QA pairs.} We construct large-scale QA pairs for IAA, IQA, and ISTA. For ISTA, we additionally design structured-output QA pairs (Fig.~\ref{fig:training_pipeline}) to support fine-grained structural reasoning.

\noindent\textbf{VR-based QA pairs.} We additionally incorporate VR data for IAA, IQA, and ISTA, also visualized in Fig.~\ref{fig:training_pipeline}, which allows the model to directly associate perceptual-level attributes with quantitative ratings.

In total, the dataset in domain-adaptive pre-training contains approximately \textbf{800K} samples. Through Domain-Adaptive Pre-Training, the model acquires the essential capability to handle diverse perceptual-level image understanding tasks across different domains.

\subsection{Task-Aligned RL for VR \& VQA}
\label{sec:rl}

\begin{table*}[t!]
\centering
\small
\caption{\textbf{Performance comparison of different models on UniPercept-Bench-VR.}
The best results are highlighted with \legendlabel{100}{dark blue} cells, and the second-best results with \legendlabel{40}{light blue} cells.
Models with * indicate those retrained on ArtiMuse-10K, KonIQ-10K, and ISTA-10K. "-/-" denotes cases where the model refuse to respond or fail to provide valid answers due to functional limitations. Metrics: SRCC/PLCC.}
\resizebox{\textwidth}{!}{
\begin{threeparttable}[t]
\renewcommand{\arraystretch}{0.9} 
\begin{tabular}{l|ccccc|ccccc|c}
\toprule[1.5pt]
\multirow{2}{*}{\textbf{Models}} &
\multicolumn{5}{c|}{\textbf{VR - IAA}} &
\multicolumn{5}{c|}{\textbf{VR - IQA}} &
\multicolumn{1}{c}{\textbf{VR - ISTA}} \\
\cmidrule(lr){2-6} \cmidrule(lr){7-11} \cmidrule(lr){12-12}
 & ArtiMuse-10K~\cite{artimuse} & AVA~\cite{ava} & TAD66K~\cite{tad66k}& FLICKR-AES~\cite{flickr} & \textbf{Avg.} &
 KonIQ-10K~\cite{koniq} & SPAQ~\cite{spaq} & KADID~\cite{kadid10k} & PIPAL~\cite{pipal}& \textbf{Avg.} &
 ISTA-10K \\
\midrule

\multicolumn{1}{l}{\textit{\textbf{Proprietary Models}}} & \multicolumn{11}{c}{\textit{\textcolor{myblue}{\textbf{Large-scale Pretraining, Cross domain testing}}}} \\
GPT-4o~\cite{gpt4}  &
0.333/0.276 & \bluecellnp{70}0.509/0.485 & 0.278/0.282 & 0.605/0.597 & \bluecellnp{70}0.431/0.410 &
0.695/0.744 & 0.874/0.881 & 0.677/0.646 & 0.325/0.349 & 0.643/0.655 & -0.003/0.116 \\
Llama-4-Scout~\cite{meta_llama4_scout_2025}  &
0.204/0.147 & 0.345/0.329 & 0.236/0.210 & 0.548/0.506 & 0.333/0.298 &
0.503/0.653 & -0.041/0.007 & -0.099/-0.004 & -0.007/0.023 & 0.089/0.170 & -0.025/0.047 \\
Gemini-2.5-pro~\cite{gemini}  &
0.187/0.035 & 0.248/0.100 & 0.143/0.037 & 0.357/0.206 & 0.234/0.095 &
0.582/0.316 & 0.087/0.212 & 0.436/0.274 & 0.225/-0.019 & 0.333/0.196 & -0.230/-0.118 \\
Claude-Sonnet-4.5~\cite{claude}  &
0.041/0.027 & 0.003/0.013 & 0.040/0.047 & 0.037/0.049 & 0.030/0.034 &
-0.037/-0.043 & 0.036/0.085 & 0.223/0.273 & -0.131/-0.088 & 0.023/0.057 & 0.125/0.089 \\
Claude-Sonnet-4.5-Think~\cite{claude}  &
0.066/0.103 & 0.018/0.019 & 0.026/0.039 & -/- & 0.037/0.054 &
-/- & -/- & -/- & -/- & -/- & -/- \\

\midrule
\multicolumn{1}{l}{\textit{\textbf{Open-Source Models}}} & \multicolumn{11}{c}{\textit{\textcolor{myblue}{\textbf{Large-scale Pretraining, Cross domain testing}}}} \\
LLaVA-OneVision-1.5-Instruct-8B~\cite{LLaVA-OneVision-1.5}  &
0.274/0.212 & 0.381/0.378 & 0.213/0.224 & 0.586/0.541 & 0.364/0.339 &
0.639/0.744 & -/- & 0.505/0.534 & 0.417/0.407 & 0.520/0.562 & -0.094/0.027 \\
GLM-4.5-V-106BA12B~\cite{glm}  &
0.346/0.249 & 0.464/0.420 & \bluecellnp{70}0.289/0.278 & \bluecellnp{70}0.651/0.597 & 0.438/0.386 &
0.721/0.765 & -0.040/-0.038 & -0.142/-0.128 & 0.013/0.020 & 0.138/0.155 & 0.083/0.117 \\
InternVL3-8B~\cite{internvl3}  &
0.245/0.211 & 0.372/0.344 & 0.205/0.191 & 0.547/0.476 & 0.342/0.306 &
0.574/0.646 & 0.828/0.800 & 0.496/0.475 & 0.435/0.459 & 0.583/0.595 & -0.127/0.046 \\
InternVL3-78B~\cite{internvl3}  &
0.223/0.206 & 0.385/0.344 & 0.221/0.220 & 0.518/0.433 & 0.337/0.301 &
0.635/0.676 & 0.849/0.852 & 0.579/0.553 & 0.415/0.457 & 0.619/0.634 & -/- \\
InternVL3.5-8B~\cite{internvl3-5}  &
0.135/0.104 & 0.308/0.295 & 0.180/0.182 & 0.519/0.448 & 0.286/0.257 &
0.663/0.660 & 0.783/0.777 & 0.541/0.478 & 0.351/0.386 & 0.585/0.575 & -0.096/-0.025 \\
InternVL3.5-38B~\cite{internvl3-5}  &
0.219/0.175 & 0.359/0.357 & 0.201/0.208 & 0.559/0.529 & 0.334/0.317 &
0.578/0.652 & 0.840/0.831 & 0.568/0.537 & 0.448/0.457 & 0.608/0.619 & \bluecellnp{70}0.262/0.345 \\
QwenVL-2.5-Instruct-7B~\cite{qwen2.5}  &
0.223/0.143 & 0.359/0.324 & 0.208/0.195 & 0.588/0.520 & 0.345/0.296 &
0.708/0.762 & -/- & 0.521/0.517 & 0.350/0.361 & 0.526/0.547 & -0.046/0.076 \\
QwenVL-2.5-Instruct-72B~\cite{qwen2.5}  &
0.233/0.197 & 0.408/0.387 & 0.232/0.235 & 0.626/0.589 & 0.375/0.352 &
0.762/0.820 & -/- & 0.606/0.570 & 0.381/0.407 & 0.583/0.599 & 0.091/0.148 \\
QwenVL-3-Instruct-8B~\cite{qwen2.5}  &
0.156/0.094 & 0.280/0.170 & 0.191/0.121 & 0.507/0.388 & 0.283/0.193 &
0.761/0.822 & 0.612/0.604 & 0.723/0.696 & 0.434/0.427 & 0.633/0.637 & 0.033/0.044 \\
QwenVL-3-Instruct-32B~\cite{qwen2.5}  &
0.227/0.130 & 0.353/0.198 & 0.200/0.095 & 0.572/0.413 & 0.338/0.209 &
0.796/0.838 & 0.690/0.657 & 0.673/0.682 & 0.414/0.402 & 0.643/0.644 & 0.084/0.106 \\

\midrule
\multicolumn{1}{l}{\textit{\textbf{Specialized Models}}} &
\multicolumn{1}{c}{\textit{\textcolor{myred}{\textbf{In domain}}}} & \multicolumn{3}{c}{\textit{\textcolor{mygreen}{\textbf{Cross domain}}}} & \multicolumn{1}{c}{\textit{\textbf{Avg.}}} &
\multicolumn{1}{c}{\textit{\textcolor{myred}{\textbf{In domain}}}} & \multicolumn{3}{c}{\textit{\textcolor{mygreen}{\textbf{Cross domain}}}} & \multicolumn{1}{c}{\textit{\textbf{Avg.}}} &
\multicolumn{1}{c}{\textit{\textcolor{myred}{\textbf{In domain}}}} \\
ArtiMuse~\cite{artimuse}  &
\bluecellnp{70}0.614/0.627 & 0.397/0.385 & 0.230/0.232 & 0.349/0.334 & 0.398/0.395 &
-/- & -/- & -/- & -/- & -/- & -/- \\
DeQA~\cite{deqa_score}  &
-/- & -/- & -/- & -/- & -/- &
\bluecellnp{100}0.953/0.941 & 0.895/0.896 & 0.694/0.687 & 0.472/0.478 & 0.753/0.750 & -/- \\
Q-Align*~\cite{qalign}   &
0.551/0.573 & 0.398/0.386 & 0.194/0.197 & 0.137/0.123 & 0.320/0.320 &
0.941/0.940 & 0.886/0.887 & 0.674/0.684 & 0.403/0.419 & 0.726/0.733 & -/- \\
Q-Insight~\cite{Q-insight}  &
-/- & -/- & -/- & -/- & -/- &
0.933/0.916 & \bluecellnp{100}0.907/0.905 & \bluecellnp{70}0.742/0.736 & \bluecellnp{70}0.486/0.474 & \bluecellnp{70}0.767/0.758 & -/- \\
Q-Insight*~\cite{Q-insight}  &
0.228/0.175 & 0.405/0.376 & 0.212/0.217 & 0.617/0.537 & 0.366/0.326 &
0.733/0.750 & 0.800/0.938 & 0.580/0.548 & 0.369/0.368 & 0.621/0.651 & 0.060/0.152 \\
\midrule
\textbf{UniPercept (Ours)}   &
\textbf{\bluecellnp{100}{0.746/0.738}} & \textbf{\bluecellnp{100}0.589/0.577} & \textbf{\bluecellnp{100}0.336/0.346} & \textbf{\bluecellnp{100}0.688/0.681} & \textbf{\bluecellnp{100}0.590/0.586} &
\textbf{\bluecellnp{70}0.940/0.949} & \textbf{\bluecellnp{70}0.904/0.895} & \textbf{\bluecellnp{100}0.872/0.870} & \textbf{\bluecellnp{100}0.581/0.594} & \textbf{\bluecellnp{100}0.824/0.827} & \textbf{\bluecellnp{100}0.778/0.767} \\
\bottomrule[1.5pt]
\end{tabular}
\end{threeparttable}}
\label{tab:vr_leaderboard}
\end{table*}

To achieve precise alignment of the model across both Visual Rating (VR) and Visual Question Answering (VQA), we employ the GRPO algorithm~\cite{grpo} to perform \textbf{Task-Aligned RL} with task-specific reward functions for VR and VQA. For the VQA task, we adopt a binary reward:
\begin{equation}
r_{vqa} =
\begin{cases}
1, & \text{if the predicted answer is correct,} \\
0, & \text{otherwise.}
\end{cases}
\label{eq:vqa_reward}
\end{equation}

For the VR task, we design an \textbf{Adaptive Gaussian Soft Reward}, which continuously evaluates the prediction according to its numerical deviation from the ground truth:
\begin{equation}
r_{vr} = \exp\!\left(-\frac{(|p_i - g_i|)^2}{2\sigma_{\text{dyn}}^2}\right), \quad 
\sigma_{\text{dyn}} = \sigma_0 \left(1 + \alpha \frac{|p_i - g_i|}{100}\right),
\label{eq:vr_reward}
\end{equation}
where $p_i$ and $g_i$ denote the predicted and ground-truth scores (mapped to $[0,100]$), $\sigma_0$ is the base smoothing coefficient, and $\alpha$ controls the degree of adaptive Gaussian smoothing. This soft reward offers smoother gradients and avoids threshold-induced discontinuities. Following prior works~\cite{qalign,artimuse}, we adopt a \textit{Token As Score} strategy for VR, deriving ratings from the predicted token distribution.
We then incorporate this task-specific reward into the GRPO objective to enable perceptual-level policy optimization:

\begin{equation}
\begin{aligned}
\mathcal{J}^{\mathcal{B}}_{\text{GRPO}}(\theta)
&= \mathbb{E}_{\mathcal{B}}\!\Bigg[
\frac{1}{\sum_{i=1}^{G} |o^i|}
\sum_{i=1}^{G} \sum_{t=1}^{|o^i|}
r_i \cdot
\min\!\Big(
r_t^i(\theta)\hat{A}_t^i, \\
&\qquad\quad
\mathrm{clip}\big(r_t^i(\theta), 1-\epsilon, 1+\epsilon\big)\hat{A}_t^i
\Big)
\Bigg].
\end{aligned}
\label{eq:grpo_objective}
\end{equation}
where $r_t^i(\theta) = \frac{\pi_\theta(o_t^i|q^i,o_{<t}^i)}{\pi_{\text{old}}(o_t^i|q^i,o_{<t}^i)}$ is the ratio between current and old policies, $\hat{A}_t^i$ is the estimated advantage, and $r_i$ is the task-specific reward from Eq.~\ref{eq:vqa_reward} or Eq.~\ref{eq:vr_reward}. 
This unified formulation enables GRPO to align model behavior with both discrete correctness (VQA) and continuous perceptual consistency (VR).

\section{Experiments}
\label{sec:experiments}

\subsection{Implementation}
\label{sec:implementation}

\noindent\textbf{Evaluated Models.}
We evaluate a total of 18 models, encompassing three categories. (1) Proprietary Models: GPT-4o~\cite{gpt4}, Llama-4-Scout~\cite{meta_llama4_scout_2025}, Gemini-2.5-Pro~\cite{gemini}, Claude-Sonnet-4.5~\cite{claude} and Claude-Sonnet-4.5-Think~\cite{claude}. (2) Leading Open-Source Models: the InternVL3 and InternVL3.5 series~\cite{internvl3, internvl3-5}, QwenVL-2.5-Instruct and QwenVL-3-Instruct series~\cite{qwen2.5}, GLM-4.5-V-106BA12B~\cite{glm}, as well as LLaVA-OneVision-1.5-Instruct~\cite{LLaVA-OneVision-1.5}. (3) Specialized Models for IAA and IQA: Q-Align~\cite{qalign}, ArtiMuse~\cite{artimuse}, DeQA~\cite{deqa_score}, and Q-Insight~\cite{Q-insight}.

\noindent\textbf{Evaluation Settings.}
For VQA, all models were provided with identical prompts corresponding to each question, and their generated answers were compared against the ground-truth options.
For VR, we designed task-specific prompts for models lacking a dedicated scoring interface to elicit quantitative predictions, while specialized models for visual rating were directly evaluated through their native interfaces. It is worth noting that all specialized models are trained exclusively on \textit{In-domain} datasets.
For representative models (Q-Align and Q-Insight), we further retrain them on a mixed dataset comprising ArtiMuse-10K~\cite{artimuse}, KonIQ-10K~\cite{koniq}, and ISTA-10K to ensure consistent perceptual alignment across domains.

\noindent\textbf{Training Details.}
Based on InternVL3-8B\cite{internvl3}, UniPercept is trained in two stages as described in Sec.~\ref{sec:baseline}. 
The Domain-Adaptive Pre-Training stage adopts multiple public datasets, including APDDv2~\cite{apddv2} and Impressions~\cite{impressions} for IAA, Q-Ground-100K~\cite{q-ground} and DataDepictQA~\cite{depictqa_v2} for IQA, and DTD~\cite{dtd}, FMD~\cite{fmd}, Flickr2K~\cite{flickr2k}, and LSDIR~\cite{lsdir} for ISTA, among others. After preprocessing and filtering, the resulting corpus contains approximately 800K samples in total.
For Task-Aligned RL, we adopt the VR datasets ArtiMuse-10K~\cite{artimuse}, KonIQ-10K~\cite{koniq}, and ISTA-10K, along with $\sim$30K VQA samples generated as described in Sec.~\ref{sec:benchmark_construction}. 
Training is performed on 16 NVIDIA A100 GPUs for 2 epochs per each stage with a batch size of 128. 
In GRPO, we sample $n{=}8$ responses per query and set $\beta{=}0.001$, $\varepsilon{=}0.2$, and $\sigma{=}0.8$.

\subsection{Benchmark Results with Analysis}
\label{sec:benchmark_results}

\subsubsection{Visual Rating}
\label{sec:benchmark_results_vr}

\begin{table*}[t]
\centering
\small
\caption{\textbf{Performance comparison of different models on UniPercept-Bench-VQA (IAA).} 
Category names are abbreviated; full definitions are provided in the Appendix. Results follow the same notation throughout the paper.}
\vspace{-6pt}
\resizebox{\textwidth}{!}{
\begin{threeparttable}
\renewcommand{\arraystretch}{0.85} 
\begin{tabular}{l|cccccccc|cccccc|c}
\toprule[1.5pt]
\multirow{2}{*}{\textbf{Models}} &
\multicolumn{8}{c|}{\textbf{IAA Categories}} &
\multicolumn{6}{c|}{\textbf{QA Templates}} & \multirow{2}{*}{\textbf{Overall}} \\
\cmidrule(lr){2-9} \cmidrule(lr){10-15} 
 & \textit{Comp.} & \textit{VisStr.} & \textit{Tech.} & \textit{Creat.} & \textit{Theme.} & \textit{Emo.} & \textit{Gest.} & \textit{CompEv.} &
\textit{Lv.Pred} & \textit{How} & \textit{What} & \textit{Which} & \textit{Why} & \textit{Yes-No} \\
\midrule
\textit{Random Guess}  &  23.08\%  &  27.27\%  &  21.95\%  &  29.63\%  &  25.93\%  &  22.86\%  &  23.68\%  &  32.56\%  &  24.14\%  &  21.28\%  &  30.43\%  &  25.32\%  &  24.00\%  &  29.49\%  &  25.17\% \\
\multicolumn{16}{l}{\textit{\textbf{Proprietary Models}}} \\
GPT-4o  &  64.62\%  &  59.57\%  &  57.58\%  &  60.19\%  &  65.19\%  &  67.62\%  &  51.95\%  &  30.23\%  &  38.86\%  &  78.17\%  &  72.46\%  &  62.66\%  &  72.67\%  &  70.51\%  &  60.04\% \\
Llama-4-Scout &  62.56\%  &  68.45\%  &  59.76\%  &  61.11\%  &  57.78\%  &  70.48\%  &  48.68\%  &  32.56\%  &  43.97\%  &  70.92\%  &  69.57\%  &  61.39\%  &  77.33\%  &  70.51\%  &  60.91\% \\
Gemini-2.5-pro  & \bluecell{70}{71.79} &  68.45\%  &  61.59\%  & \bluecell{70}{76.85} &  67.41\%  &  63.81\%  & \bluecell{70}{61.84} &  37.21\%  &  45.98\%  &  78.72\%  &  73.91\%  &  67.72\%  &  84.67\%  &  \bluecell{100}{84.62} &  66.44\% \\
Claude-Sonnet-4.5  &  70.26\%  &  70.05\%  &  62.20\%  &  71.30\%  &  64.44\%  &  67.62\%  &  50.00\%  & \bluecell{70}{46.51} & \bluecell{70}{46.84} &  77.30\%  &  76.09\%  &  65.19\%  &  86.00\%  &  69.23\%  &  65.45\% \\
Claude-Sonnet-4.5-Think  &  71.28\%  &  69.52\%  &  61.21\%  &  68.52\%  &  62.22\%  &  66.67\%  &  53.25\%  &  41.86\%  &  44.57\%  &  75.89\%  &  77.54\%  &  67.09\%  &  86.00\%  &  66.67\%  &  64.73\% \\
\midrule
\multicolumn{16}{l}{\textit{\textbf{Open-Source Models}}} \\
LLaVA-OneVision-1.5-Instruct-8B  &  67.18\%  &  68.62\%  &  61.21\%  &  62.96\%  &  67.41\%  &  62.86\%  &  53.25\%  &  20.93\%  &  34.86\%  &  85.21\%  &  79.71\%  &  65.82\%  &  83.33\%  &  69.23\%  &  62.60\% \\
GLM-4.5-V-106BA12B  &  67.18\%  &  65.78\%  &  60.98\%  &  75.00\%  &  64.44\%  &  68.57\%  &  51.32\%  &  \bluecell{70}{46.51} &  45.40\%  &  71.63\%  &  78.26\%  &  65.82\%  &  84.67\%  &  70.51\%  &  64.46\% \\
InternVL3-8B  &  65.64\%  &  67.55\%  &  59.39\%  &  67.59\%  &  69.63\%  &  62.86\%  &  50.65\%  &  25.58\%  &  36.00\%  &  81.69\%  &  73.91\%  &  67.72\%  &  86.00\%  &  71.79\%  &  62.60\% \\
InternVL3-78B  & \bluecell{70}{71.79} & \bluecell{70}{73.26} &  61.21\%  &  73.15\%  & \bluecell{70}{74.81} & \bluecell{70}{74.29} &  53.25\%  &  37.21\%  &  45.14\%  &  85.82\%  & \bluecell{70}{81.16} &  72.15\%  &  86.00\%  &  75.64\%  & \bluecell{70}{68.28} \\
InternVL3.5-8B  &  32.31\%  &  29.41\%  &  30.30\%  &  26.85\%  &  28.89\%  &  26.67\%  &  23.38\%  &  9.30\%  &  17.14\%  &  41.13\%  &  26.81\%  &  19.62\%  &  36.00\%  &  58.97\%  &  28.18\% \\
InternVL3.5-38B  &  37.44\%  &  40.11\%  &  27.88\%  &  39.81\%  &  34.81\%  &  38.10\%  &  45.45\%  &  6.98\%  &  34.00\%  &  47.52\%  &  26.09\%  &  28.48\%  &  37.33\%  &  50.00\%  &  35.67\% \\
QwenVL-2.5-Instruct-7B  &  67.18\%  &  70.74\%  &  56.36\%  &  66.67\%  &  68.89\%  &  63.81\%  &  48.05\%  &  37.21\%  &  38.86\%  &  76.76\%  &  75.36\%  &  67.09\%  &  \bluecell{100}{87.33} &  71.79\%  &  63.19\% \\
QwenVL-2.5-Instruct-72B  &  22.05\%  &  24.60\%  &  25.45\%  &  29.63\%  &  30.37\%  &  18.10\%  &  19.48\%  &  6.98\%  &  14.00\%  &  19.86\%  &  17.39\%  &  24.05\%  &  41.33\%  &  51.28\%  &  23.74\% \\
QwenVL-3-Instruct-8B  &  31.28\%  &  32.09\%  &  32.12\%  &  37.04\%  &  34.07\%  &  22.86\%  &  37.66\%  &  25.58\%  &  35.43\%  &  14.89\%  &  17.39\%  &  34.81\%  &  28.67\%  &  73.08\%  &  31.92\% \\
QwenVL-3-Instruct-32B  &  23.08\%  &  26.74\%  &  32.12\%  &  26.85\%  &  32.59\%  &  20.95\%  &  33.77\%  &  20.93\%  &  33.43\%  &  9.22\%  &  13.77\%  &  31.01\%  &  18.67\%  &  66.67\%  &  27.39\% \\
\midrule
\multicolumn{16}{l}{\textit{\textbf{Specialized Models}}} \\
ArtiMuse  &  67.69\%  &  68.45\%  &  \bluecell{70}{64.85}  &  74.07\%  &  71.85\%  &  64.76\%  &  61.04\%  &  32.56\%  &  39.14\%  & \bluecell{70}{88.65} &  76.81\%  & \bluecell{70}{72.78} &  85.33\%  &  \bluecell{70}{79.49}  &  66.31\% \\
\midrule
\textbf{UniPercept (Ours)}  & \textbf{\bluecell{100}{80.00}} & \textbf{\bluecell{100}{77.54}} & \textbf{\bluecell{100}{69.70}} & \textbf{\bluecell{100}{80.56}} & \textbf{\bluecell{100}{79.26}} & \textbf{\bluecell{100}{80.95}} & \textbf{\bluecell{100}{67.53}} & \textbf{\bluecell{100}{69.77}} & \textbf{\bluecell{100}{63.71}} & \textbf{\bluecell{100}{92.20}} & \textbf{\bluecell{100}{81.88}} & \textbf{\bluecell{100}{75.32}} & \textbf{\bluecell{70}{86.67}} & \textbf{\bluecell{100}{84.62}} & \textbf{\bluecell{100}{76.55}} \\
\bottomrule[1.5pt]
\end{tabular}
\end{threeparttable}}
\label{tab:vqa_iaa_leaderboard}
\end{table*}

\begin{figure*}[htbp]
\centering
\begin{minipage}[t]{0.72\textwidth}  
    \vspace{0pt}  
    \centering
    \small
    \captionof{table}{\textbf{Performance comparison of different models on UniPercept-Bench-VQA (IQA).}}
    \vspace{-6pt}
    \resizebox{\textwidth}{!}{%
    \begin{threeparttable}
\renewcommand{\arraystretch}{0.9} 
\begin{tabular}{l|ccc|cccccc|c}
\toprule[1.5pt]
\multirow{2}{*}{\textbf{Models}} &
\multicolumn{3}{c|}{\textbf{IQA Categories}} &
\multicolumn{6}{c|}{\textbf{QA Templates}} &
\multirow{2}{*}{\textbf{Overall}} \\
\cmidrule(lr){2-4} \cmidrule(lr){5-10}
 & \textit{Loc.} & \textit{Sev.} & \textit{Type.} & \textit{Lv.Pred} & \textit{How} & \textit{What} & \textit{Which} & \textit{Why} & \textit{Yes-No} \\
\midrule
\textit{Random Guess}   &
23.67\%  &  24.75\%  &  20.08\%  &  24.75\%  &
27.03\%  &  16.05\%  &  25.00\%  &  21.39\%  &
22.99\%  &  23.16\% \\
\multicolumn{11}{l}{\textit{\textbf{Proprietary Models}}} \\
GPT-4o   &
71.74\%  &  53.18\%  &  70.49\%  &  53.18\%  &
83.78\%  &  59.26\%  &  61.31\%  &  80.21\%  &
67.82\%  &  66.36\% \\
Llama-4-Scout   &
60.18\%  &  58.19\%  &  52.05\%  &  58.19\%  &
82.16\%  &  37.04\%  &  38.69\%  &  66.31\%  &
62.07\%  &  57.81\% \\
Gemini-2.5-pro   &
32.84\%  &  52.84\%  &  40.98\%  &  52.84\%  &
40.54\%  &  32.72\%  &  29.17\%  &  41.18\%  &
28.74\%  &  40.17\% \\
Claude-Sonnet-4.5   &
71.19\%  &  51.51\%  &  66.80\%  &  51.51\%  &
90.81\%  &  50.00\%  &  50.60\%  &  82.89\%  &
71.26\%  &  64.80\% \\
Claude-Sonnet-4.5-Think   &
71.19\%  &  55.52\%  &  66.80\%  &  55.52\%  &
89.19\%  &  50.00\%  &  51.79\%  &  82.89\%  &
72.41\%  &  65.90\% \\
\midrule
\multicolumn{11}{l}{\textit{\textbf{Open-Source Models}}} \\
LLaVA-OneVision-1.5-Instruct-8B   &
\bluecell{70}{76.51}  &  \bluecell{70}{59.87}  &  77.46\%  &  \bluecell{70}{59.87}  &
\bluecell{70}{91.35}  &  \bluecell{70}{70.37}  &  61.31\%  &  82.35\%  &
75.86\%  &  \bluecell{70}{72.15} \\
GLM-4.5-V-106BA12B   &
70.09\%  &  35.79\%  &  54.51\%  &  35.79\%  &
88.11\%  &  48.77\%  &  44.05\%  &  74.33\%  &
68.97\%  &  57.17\% \\
InternVL3-8B   &
71.56\%  &  52.84\%  &  59.43\%  &  52.84\%  &
87.03\%  &  59.88\%  &  48.81\%  &  71.12\%  &
71.26\%  &  63.69\% \\
InternVL3-78B   &
75.41\%  &  51.84\%  &  \bluecell{70}{81.56}  &  51.84\%  &
\bluecell{100}{93.51}  &  66.67\%  &  \bluecell{70}{63.10}  &  \bluecell{100}{88.24}  &
66.67\%  &  70.31\% \\
InternVL3.5-8B   &
38.17\%  &  44.82\%  &  38.11\%  &  44.82\%  &
35.14\%  &  41.98\%  &  30.36\%  &  36.36\%  &
56.32\%  &  39.98\% \\
InternVL3.5-38B   &
38.90\%  &  49.83\%  &  45.08\%  &  49.83\%  &
46.49\%  &  41.36\%  &  31.55\%  &  33.16\%  &
62.07\%  &  43.29\% \\
QwenVL-2.5-Instruct-7B   &
74.13\%  &  48.83\%  &  66.39\%  &  48.83\%  &
88.65\%  &  60.49\%  &  53.57\%  &  78.61\%  &
\bluecell{70}{77.01}  &  65.44\% \\
QwenVL-2.5-Instruct-72B   &
31.01\%  &  4.68\%  &  16.39\%  &  4.68\%  &
35.14\%  &  14.81\%  &  11.31\%  &  22.99\%  &
66.67\%  &  20.50\% \\
QwenVL-3-Instruct-8B   &
34.68\%  &  55.18\%  &  16.39\%  &  55.18\%  &
20.54\%  &  18.52\%  &  27.38\%  &  25.67\%  &
\bluecell{70}{77.01}  &  36.21\% \\
QwenVL-3-Instruct-32B   &
29.54\%  &  14.38\%  &  16.80\%  &  14.38\%  &
11.89\%  &  18.52\%  &  25.60\%  &  22.46\%  &
74.71\%  &  22.52\% \\
\midrule
\textbf{UniPercept (Ours)}   &
\textbf{\bluecell{100}{77.43}}  &  \textbf{\bluecell{100}{79.60}}  &  \textbf{\bluecell{100}{90.98}}  &  \textbf{\bluecell{100}{79.60}}  &
\textbf{87.03\%}  &  \textbf{\bluecell{100}{80.86}}  &  \textbf{\bluecell{100}{75.60}}  &  \textbf{\bluecell{70}{83.42}}  &
\textbf{\bluecell{100}{79.31}}  &  \textbf{\bluecell{100}{81.07}} \\
\bottomrule[1.5pt]
\end{tabular}
\end{threeparttable}}
\label{tab:vqa_iqa_leaderboard}
\end{minipage}
\hfill
\begin{minipage}[t]{0.27\textwidth}  
    \vspace{0pt}  
    \centering
    \includegraphics[width=\linewidth]{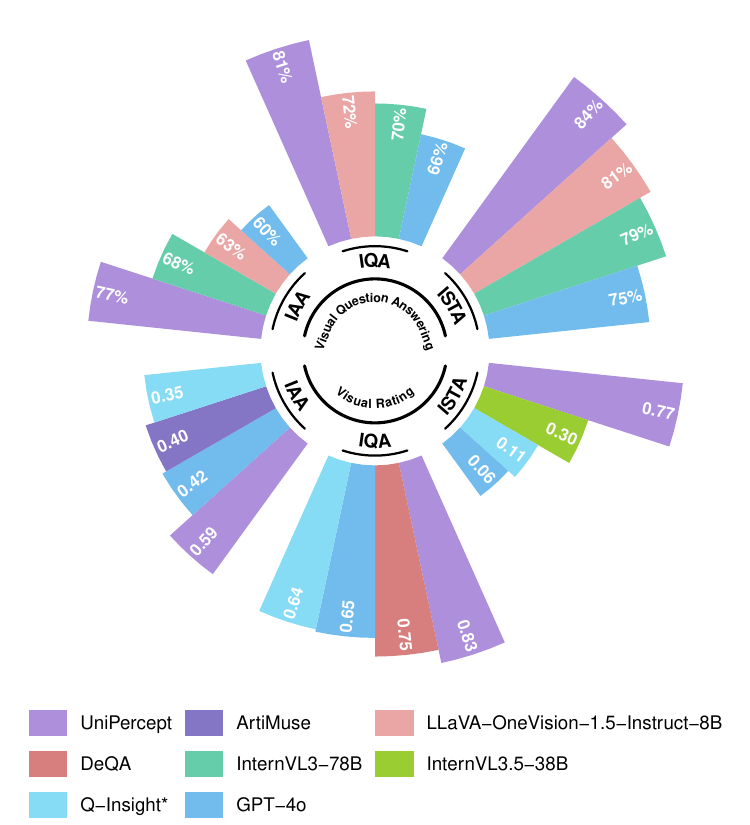}
    \vspace{-20pt}
    \caption{\textbf{Results on UniPercept-Bench.} 
    VQA and VR are evaluated by Acc. and $(SRCC+PLCC)/2$.}
    \label{fig:benchmark_results_radar}
\end{minipage}
\end{figure*}

\begin{table*}[htbp]
\centering
\scriptsize  
\caption{\textbf{Performance comparison of different models on UniPercept-Bench-VQA (ISTA).}}
\vspace{-6pt}
\label{tab:vqa_ista_leaderboard}

\begin{threeparttable}
\renewcommand{\arraystretch}{0.75}   
\setlength{\tabcolsep}{2pt}         

\begin{tabular*}{\textwidth}{@{\extracolsep{\fill}} l|ccccc|ccccc|c}
\toprule[1.5pt]
\multirow{2}{*}{\textbf{Models}} &
\multicolumn{5}{c|}{\textbf{ISTA Categories}} &
\multicolumn{5}{c|}{\textbf{QA Templates}} &
\multirow{2}{*}{\textbf{Overall}} \\
\cmidrule(lr){2-6} \cmidrule(lr){7-11}
 & \textit{Scene.} & \textit{Phys.} & \textit{Mat.} & \textit{Geo.} & \textit{Sem.} &
   \textit{How} & \textit{What} & \textit{Which} & \textit{Why} & \textit{Yes-No} \\
\midrule
\textit{Random Guess}   & 26.50\% & 23.63\% & 24.73\% & 30.30\% & 30.58\% & 26.28\% & 23.84\% & 24.29\% & 33.77\% & 33.33\% & 26.60\% \\
\multicolumn{12}{l}{\textit{\textbf{Proprietary Models}}} \\
GPT-4o                 & 75.64\% & 79.12\% & 73.48\% & 33.33\% & 77.27\% & 71.79\% & 78.78\% & 69.23\% & 77.92\% & 72.46\% & 74.64\% \\
Llama-4-Scout          & 73.50\% & 75.27\% & 71.68\% & 72.73\% & 67.77\% & 75.64\% & 69.77\% & 69.64\% & 77.27\% & 69.57\% & 71.86\% \\
Gemini-2.5-pro         & 76.50\% & 82.42\% & 77.06\% & 66.67\% & 77.69\% & 78.21\% & 78.20\% & 75.71\% & 82.47\% & 71.01\% & 77.73\% \\
Claude-Sonnet-4.5      & 76.92\% & 78.57\% & 74.91\% & \bluecell{70}{90.91} & 77.69\% &
                         76.92\% & 77.03\% & 74.49\% & 81.82\% & 79.71\% & 77.32\% \\
Claude-Sonnet-4.5-Think& 77.35\% & 78.02\% & 73.12\% & 87.88\% & 75.21\% &
                         76.28\% & 74.71\% & 74.09\% & 81.82\% & 76.81\% & 76.08\% \\
\midrule
\multicolumn{12}{l}{\textit{\textbf{Open-Source Models}}} \\
LLaVA-OneVision-1.5-Instruct-8B & 78.63\% & \bluecell{70}{85.16} & \bluecell{100}{82.44} & 72.73\% & \bluecell{100}{80.17} &
                                   \bluecell{100}{83.33} & \bluecell{70}{81.40} & 75.30\% & \bluecell{100}{84.42} & \bluecell{100}{88.41} & \bluecell{70}{81.13} \\
GLM-4.5-V-106BA12B              & \bluecell{70}{81.20} & 79.67\% & 74.55\% & 72.73\% & 75.21\% &
                                   80.77\% & 76.74\% & 73.68\% & 79.87\% & 78.26\% & 77.22\% \\
InternVL3-8B                    & 75.64\% & 79.12\% & 73.48\% & 33.33\% & 77.27\% &
                                   71.79\% & 78.78\% & 69.23\% & 77.92\% & 72.46\% & 74.64\% \\
InternVL3-78B                   & 79.06\% & \bluecell{70}{85.16} & \bluecell{70}{77.42} & 69.70\% & \bluecell{70}{78.51} &
                                   81.41\% & 79.65\% & 73.68\% & \bluecell{100}{84.42} & 81.16\% & 79.28\% \\
InternVL3.5-8B                  & 54.27\% & 50.55\% & 58.42\% & 39.39\% & 36.36\% &
                                   46.79\% & 56.69\% & 48.58\% & 29.87\% & 71.01\% & 49.79\% \\
InternVL3.5-38B                 & 50.00\% & 55.49\% & 61.29\% & 30.30\% & 35.95\% &
                                   50.64\% & 59.30\% & 42.91\% & 37.01\% & 57.97\% & 50.10\% \\
QwenVL-2.5-Instruct-7B          & 74.79\% & 72.53\% & 74.91\% & 51.52\% & 73.55\% &
                                   73.72\% & 77.33\% & 66.80\% & 74.03\% & 73.91\% & 73.30\% \\
QwenVL-2.5-Instruct-72B         & 14.10\% & 29.12\% & 19.71\% & 12.12\% & 18.60\% &
                                   20.51\% & 12.21\% & 14.57\% & 31.17\% & 46.38\% & 19.59\% \\
QwenVL-3-Instruct-8B            & 27.78\% & 32.42\% & 25.45\% & 39.39\% & 24.79\% &
                                   14.74\% & 23.26\% & 28.34\% & 25.32\% & 81.16\% & 27.63\% \\
QwenVL-3-Instruct-32B           & 26.50\% & 24.73\% & 19.00\% & 15.15\% & 18.60\% &
                                   11.54\% & 18.31\% & 22.67\% & 17.53\% & 66.67\% & 21.65\% \\
\midrule
\rowcolor{tableblue!8}  
\textbf{UniPercept (Ours)} &
  \textbf{\bluecell{100}{89.74}} &
  \textbf{\bluecell{100}{85.71}} &
  \textbf{\bluecell{100}{82.44}} &
  \textbf{\bluecell{100}{93.94}} &
  \textbf{\bluecell{70}{78.51}}  &
  \textbf{\bluecell{70}{82.69}}  &
  \textbf{\bluecell{100}{89.24}} &
  \textbf{\bluecell{100}{78.54}} &
  \textbf{\bluecell{70}{83.12}}  &
  \textbf{\bluecell{70}{85.51}}  &
  \textbf{\bluecell{100}{84.23}} \\
\bottomrule[1.5pt]
\end{tabular*}
\end{threeparttable}

\end{table*}

\noindent\textbf{General Models vs. Specialized Models.}
Visual Rating is a challenging task that requires models to output continuous, high-precision perceptual scores. 
As shown in Table~\ref{tab:vr_leaderboard}, most \textit{general-purpose MLLMs} without task-specific training exhibit significantly lower performance compared with \textit{specialized models}. 
This gap arises from the inherent difficulty of directly generating numerical outputs in text form, which often leads to hallucinated or unstable predictions---a phenomenon also discussed in prior works~\cite{qalign,nexttoken,artimuse}. 
In contrast, specialized methods adopt structured strategies such as \textit{Token As Score}, effectively stabilizing the regression behavior of large models on continuous scales.~\cite{qalign,artimuse}

\noindent\textbf{Domain-wise Analysis.}
Model performance varies significantly across different domains, reflecting differences in task objectivity. 
IQA achieves the highest correlations, followed by ISTA, while IAA remains the most challenging due to its subjective and ambiguous aesthetic judgments. 
These results suggest that perceptual-level visual rating becomes increasingly difficult as the task shifts from objective assessment toward subjective reasoning.

\subsubsection{Visual Question Answering}
\label{sec:benchmark_results_vqa}

\noindent\textbf{Domain-wise Analysis.}
Across domains, the best-performing generalized models can reach accuracies of up to 68.28\%, 72.15\%, and 81.13\% on IAA, IQA, and ISTA, respectively. 
However, their average accuracies drop to only \textbf{51.62\%}, \textbf{51.45\%}, and \textbf{60.56\%}. 
This disparity suggests that perceptual-level visual question answering remains highly challenging for current MLLMs, as they often rely on unstable domain-specific heuristics rather than robust perceptual reasoning. 
Among the three domains, ISTA appears slightly easier, as it centers on objective and physically grounded properties—such as geometry, structure, and material composition—that align better with the visual–text priors learned during large-scale pretraining. In contrast, IAA and IQA require subjective reasoning over aesthetics, emotion, and perceptual quality—nuanced judgments that general models are not explicitly optimized for.

\noindent\textbf{Category-wise Analysis.}
Models generally perform better on holistic perception categories such as \textit{Composition \& Design} and \textit{Theme \& Communication}, which focus on the overall aesthetic and semantic coherence of an image. 
Most models achieve accuracies above \textbf{60\%} in these categories, indicating that global scene perception is relatively well captured by vision–language pretraining. 
In contrast, performance drops substantially on fine-grained perceptual categories such as \textit{Overall Gestalt}, \textit{Material Representation}, and \textit{Geometric Composition}, where most models remain below \textbf{50\%}. 
These results suggest that while high-level holistic understanding is well learned, precise reasoning over local structures cues remains a key limitation.

\noindent\textbf{QA Template Analysis.}
The \textit{Level Prediction} questions require models to provide fine-grained evaluations of images along specific dimensions (as shown in Fig.~\ref{fig:qa_example}). 
Due to the demand for precise quantitative reasoning, most MLLMs struggle with this type, achieving only around \textbf{36\%} and \textbf{45\%} average accuracy on the IAA and IQA domains, respectively. 
For other QA templates, “\textit{Yes-No}” and “\textit{Why}” questions—closer to higher-level reasoning and causal inference—allow generalized MLLMs to reach average accuracies above \textbf{60\%}. 
In contrast, “\textit{What}” and “\textit{Which}” questions, which require detailed and localized visual analysis, remain challenging, reflecting the current models’ limited capacity for fine-grained perceptual understanding.

\subsection{Further Discussion on UniPercept}
\subsubsection{Performance}
\label{sec:performance_unipercept}

As shown in Fig.~\ref{fig:benchmark_results_radar} and Tabs.~\ref{tab:vr_leaderboard},~\ref{tab:vqa_iaa_leaderboard}, ~\ref{tab:vqa_iqa_leaderboard}, and  ~\ref{tab:vqa_ista_leaderboard}, UniPercept is capable of handling both major types of perceptual-level image understanding tasks (\textit{VR} \& \textit{VQA}) across the three perceptual domains of \textit{IAA}, \textit{IQA}, and \textit{ISTA}. 
Benefiting from the proposed \textbf{Domain-Adaptive Pre-Training} and \textbf{Task-Aligned RL} strategies, UniPercept consistently outperforms both generalized and specialized models. 
It not only achieves strong in-domain results but also exhibits remarkable cross-domain generalization, demonstrating its capability as a unified perceptual understanding baseline.

\begin{table*}[t]
\vspace{-10pt}
\centering
\small
\setlength{\tabcolsep}{3pt}
\renewcommand{\arraystretch}{0.9}
\caption{\textbf{More Performance Results of FLUX.1-dev w/ UniPercept Reward.}}
\label{tab:application}
\vspace{-6pt}

\resizebox{0.99\textwidth}{!}{
\begin{tabular}{l cc c cc ccc}
\toprule
\multirow{2}{*}{\textbf{Models}} &
\multicolumn{2}{c}{\textbf{Preference Score}} &
\multicolumn{1}{c}{\textbf{Image Quality}} &
\multicolumn{2}{c}{\textbf{Image Aesthetics}} &
\multicolumn{3}{c}{\textbf{UniPercept Score}} \\
\cmidrule(lr){2-3} \cmidrule(lr){4-4}
\cmidrule(lr){5-6} \cmidrule(lr){7-9}
& PickScore~\cite{pickscore} & HPSv3~\cite{hspv3} & DeQA~\cite{deqa_score} & LAION-Aes~\cite{laion-aes} & ArtiMuse~\cite{artimuse} & IAA & IQA & ISTA \\
\midrule
Baseline 
& 22.46 & 10.71 & 4.32 & 5.77  & 59.02 & 65.18 & 73.59 & 46.64 \\

w/ UniPercept IAA Reward 
& 22.47 & 10.09 & 4.09 & \bluecellnp{100}{6.19} & \bluecellnp{100}{67.02} & \bluecellnp{100}{76.20} & 76.39 & 54.83 \\

w/ UniPercept IQA Reward  
& 22.63 & \bluecellnp{100}{11.21} & \bluecellnp{100}{4.37} & 6.02 & 63.64 & 72.16 & \bluecellnp{70}{76.87} & 52.34 \\

w/ UniPercept ISTA Reward  
& \bluecellnp{100}{22.72} & \bluecellnp{70}{11.09} & \bluecellnp{100}{4.37} & \bluecellnp{70}{6.16} & 63.75 & 72.23 & 76.17 & \bluecellnp{100}{59.61} \\

\textbf{w/ UniPercept Reward (All)} 
& \textbf{\bluecellnp{70}{22.67}} 
& \textbf{10.93} 
& \textbf{\bluecellnp{70}{4.33}} 
& \textbf{\bluecellnp{100}{6.19}} 
& \textbf{\bluecellnp{70}{65.52}} 
& \textbf{\bluecellnp{70}{74.24}} 
& \textbf{\bluecellnp{100}{77.04}} 
& \textbf{\bluecellnp{70}{59.08}} \\
\bottomrule
\end{tabular}
}

\end{table*}

\begin{figure*}[t]
    \centering
    \includegraphics[width=0.99\textwidth]{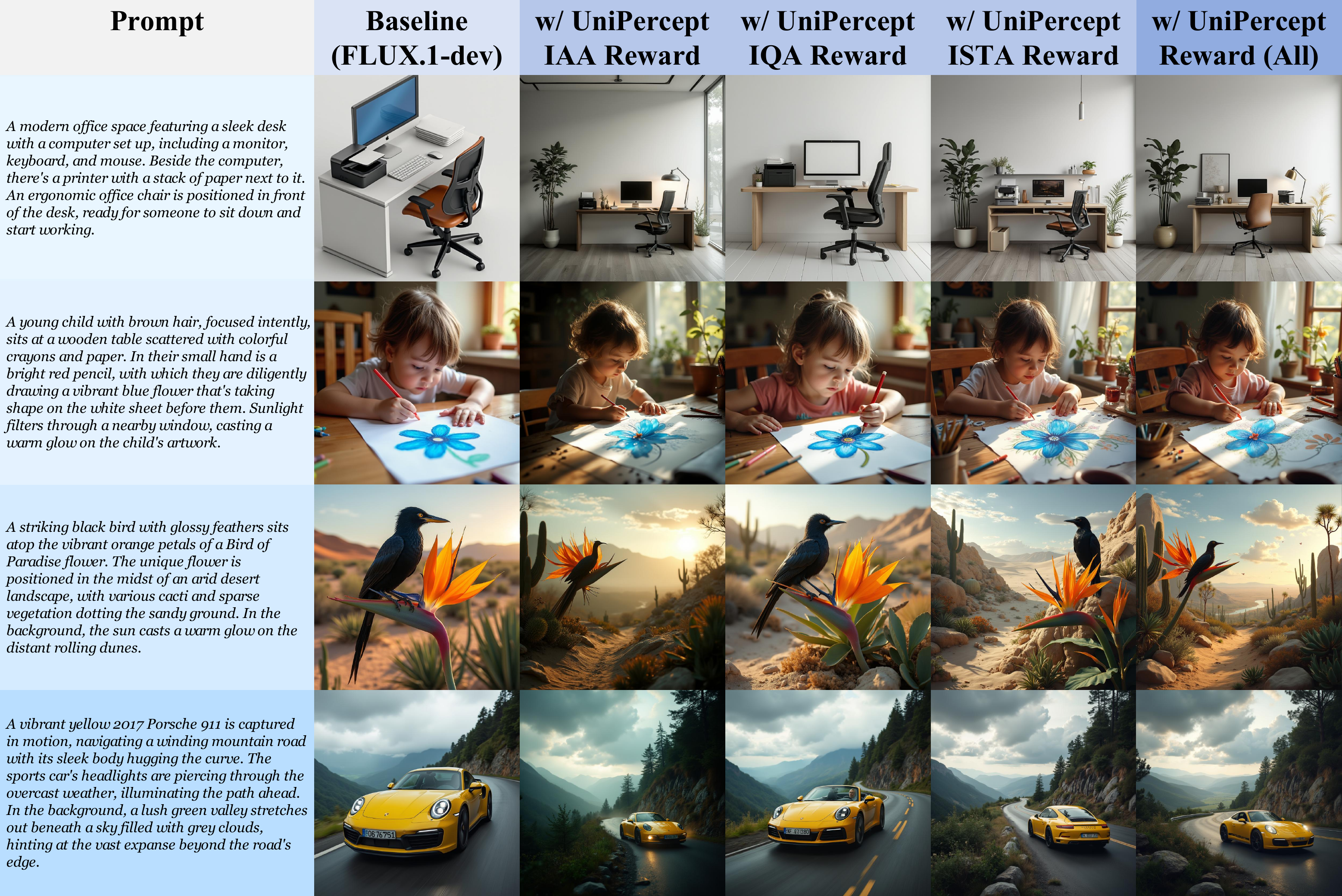}
    \vspace{-6pt}
    \caption{\textbf{More results of FLUX.1-dev w/ UniPercept Reward.} Different reward signals emphasize distinct perceptual attributes, while \textit{UniPercept Reward (All)} achieves the best overall performance by integrating complementary perceptual cues.}
    \label{fig:application_appendix}
\end{figure*}

\subsubsection{UniPercept Reward}
\label{sec:result_reward}

We investigate the use of UniPercept as a \textbf{multi-dimensional perceptual reward model} for text-to-image generation. UniPercept provides three types of visual ratings (Aesthetics Quality (IAA), Image Quality (IQA), and Structure \& Texture Richness (ISTA)) which can be employed as independent reward signals or jointly as an integrated perceptual reward. All reward configurations are incorporated into the Flow-GRPO~\cite{liu2025flowgrpotrainingflowmatching} fine-tuning pipeline based on FLUX.1-dev~\cite{flux}. Quantitative results in Tab.~\ref{tab:application} show that each reward captures a distinct facet of perceptual evaluation and thus leads to different characteristic improvements. The IAA reward mainly enhances aesthetics-related metrics, whereas the IQA reward yields stronger gains on quality-oriented metrics such as sharpness and clarity. The ISTA reward, in contrast, provides more balanced improvements across all dimensions due to its emphasis on structural and textural richness. When combining all three as a unified UniPercept reward (All), the model integrates the benefits of each dimension and achieves the best overall performance.

Corresponding qualitative results in Fig.~\ref{fig:application_appendix} further demonstrate that UniPercept-guided training improves visual fidelity, perceptual quality, and human preference. These results confirm that UniPercept serves effectively not only as a reward model but also as a unified set of perceptual metrics for evaluating generated images.

\subsubsection{UniPercept Metrics}
\label{sec:metrics}

\noindent\textbf{For Models.}
UniPercept can serve as an perceptual‐level metric that assesses the quality of outputs from any model producing images, covering three complementary dimensions: IAA, IQA, and ISTA. Here, we take text-to-image models as an example and apply UniPercept as an evaluation metric on both the DPG~\cite{dpg} and GenEval~\cite{geneval} benchmarks to measure the perceptual quality of generated images. Except for special cases (e.g., GPT-Image-1~\cite{gpt4}, where output resolution cannot be controlled), all models are evaluated with a fixed output resolution of 1024×1024. The results are reported in Tab.\ref{tab:dpg_unipercept_full} and Tab.\ref{tab:geneval_unipercept_full}.

We observe that while current models generally achieve strong performance in IQA, there remains \textbf{substantial room for improvement in IAA and ISTA}. Moreover, as model capability advances, we see consistent improvements in both instruction-following ability (as reflected by DPG and GenEval metrics) and perceptual image quality (as reflected by UniPercept). For instance, GPT-Image-1~\cite{gpt4} and Qwen-Image~\cite{wu2025qwenimagetechnicalreport} exhibit strong performance across both Dataset Metrics and UniPercept Metrics. More detailed analyses will be presented in future work.

\begin{table*}[t]
\centering
\small
\caption{\textbf{Evaluation of T2I Models on DPG~\cite{dpg} Metrics and UniPercept Metrics.}}
\vspace{-6pt}
\resizebox{0.95\textwidth}{!}{
\begin{threeparttable}[t]
\renewcommand{\arraystretch}{1.0}

\begin{tabular}{l|cccccc|cccc}
\toprule[1.5pt]
\multirow{2}{*}{\textbf{Models}} &
\multicolumn{6}{c|}{\textbf{DPG Metrics~\cite{dpg}}} &
\multicolumn{4}{c}{\textbf{UniPercept Metrics}} \\
\cmidrule(lr){2-7} \cmidrule(lr){8-11}
 & \textbf{Global} & \textbf{Entity} & \textbf{Attribute} & \textbf{Relation} & \textbf{Other} & \textbf{Overall} 
 & \textbf{IAA} & \textbf{IQA} & \textbf{ISTA} & \textbf{Avg.} \\
\midrule

OmniGen~\cite{OmniGen}                
& -- & -- & -- & -- & -- & -- 
& 62.83 & 72.22 & 45.09 & 60.04 \\

OmniGen2~\cite{OmniGen2}               
& 88.81 & 88.83 & 90.18 & 89.37 & 90.27 & 83.57
& 58.51 & 71.89 & 43.31 & 57.90 \\

BAGEL~\cite{bagel}                  
& \bluecellnp{70}{88.94} & 90.37 & \bluecellnp{70}{91.29} & 90.82 & 88.67 & 85.07
& 60.20 & 70.52 & 45.78 & 58.83 \\

SANA-1.6B~\cite{xie2024sana,xie2025sana}              
& 86.00 & 91.50 & 88.90 & 91.90 & 90.70 & 84.80
& 40.33 & 42.89 & 42.41 & 41.87 \\

Lumina-DiMOO~\cite{dimoo}           
& 81.46 & \bluecellnp{100}{92.08} & 88.98 & \bluecellnp{100}{94.31} & 82.00 & \bluecellnp{70}{86.04}
& 61.00 & 71.14 & 44.83 & 58.99 \\

FLUX.1-dev~\cite{flux}             
& 74.35 & 90.00 & 88.96 & 90.87 & 88.33 & 83.84
& \bluecellnp{100}{65.18} & \bluecellnp{100}{73.59} & \bluecellnp{70}{46.64} & \bluecellnp{100}{61.80} \\


GPT-Image-1~\cite{gpt4}            
& 88.89 & 88.94 & 89.84 & \bluecellnp{70}{92.63} & \bluecellnp{70}{90.96} & 85.15
& 62.27 & \bluecellnp{100}{72.87} & 44.88 & 60.00 \\

Qwen-Image~\cite{wu2025qwenimagetechnicalreport}             
& \bluecellnp{100}{91.32} & \bluecellnp{70}{91.56} & \bluecellnp{100}{92.02} & \bluecellnp{100}{94.31} & \bluecellnp{100}{92.73} & \bluecellnp{100}{88.32}
& \bluecellnp{70}{62.89} & 72.15 & \bluecellnp{100}{47.23} & \bluecellnp{70}{60.76} \\

\bottomrule[1.5pt]
\end{tabular}

\end{threeparttable}}
\label{tab:dpg_unipercept_full}
\end{table*}

\begin{table*}[htbp]
\centering
\small
\caption{\textbf{Evaluation of T2I Models on GenEval~\cite{geneval} Metrics and UniPercept Metrics.}}
\vspace{-6pt}
\resizebox{0.95\textwidth}{!}{
\begin{threeparttable}[t]
\renewcommand{\arraystretch}{1.0}

\begin{tabular}{l|ccccccc|cccc}
\toprule[1.5pt]
\multirow{2}{*}{\textbf{Models}} &
\multicolumn{7}{c|}{\textbf{GenEval~\cite{geneval} Metrics}} &
\multicolumn{4}{c}{\textbf{UniPercept Metrics}} \\
\cmidrule(lr){2-8} \cmidrule(lr){9-12}
& \textbf{Single Obj.} & \textbf{Two Obj.} & \textbf{Counting} & \textbf{Colors} & \textbf{Position} & \textbf{Attr. Bind.} & \textbf{Overall}
& \textbf{IAA} & \textbf{IQA} & \textbf{ISTA} & \textbf{Avg.} \\
\midrule

OmniGen~\cite{OmniGen}
& \bluecellnp{70}{0.99} & 0.86 & 0.64 & 0.85 & 0.31 & 0.55 & 0.70
& 58.84 & \bluecellnp{70}{75.62} & 41.00 & 58.49 \\

OmniGen2~\cite{OmniGen2}
& \bluecellnp{70}{0.99} & \bluecellnp{100}{0.96} & 0.74 & \bluecellnp{100}{0.98} & 0.71 & 0.75 & 0.86
& 54.20 & 75.16 & 34.48 & 54.61 \\

BAGEL~\cite{bagel}
& \bluecellnp{70}{0.99} & \bluecellnp{70}{0.94} & 0.81 & 0.88 & 0.64 & 0.63 & 0.82
& 58.68 & 71.24 & 38.35 & 56.09 \\

SANA-1.6B~\cite{xie2024sana,xie2025sana}
& \bluecellnp{70}{0.99} & 0.77 & 0.62 & 0.88 & 0.21 & 0.47 & 0.66
& 34.34 & 35.11 & 31.22 & 33.56 \\

Lumina-DiMOO~\cite{dimoo}
& \bluecellnp{100}{1.00} & \bluecellnp{70}{0.94} & \bluecellnp{70}{0.85} & 0.89 & \bluecellnp{100}{0.85} & \bluecellnp{70}{0.76} & \bluecellnp{100}{0.88}
& 51.93 & 71.98 & 30.86 & 51.59 \\

FLUX.1-dev~\cite{flux}
& 0.98 & 0.81 & 0.74 & 0.79 & 0.22 & 0.45 & 0.66
& \bluecellnp{70}{64.24} & 74.96 & \bluecellnp{70}{41.14} & \bluecellnp{70}{60.11} \\


GPT-Image-1~\cite{gpt4}
& \bluecellnp{70}{0.99} & 0.92 & \bluecellnp{70}{0.85} & \bluecellnp{70}{0.92} & 0.75 & 0.61 & 0.84
& \bluecellnp{100}{69.07} & \bluecellnp{100}{76.74} & \bluecellnp{100}{51.26} & \bluecellnp{100}{65.69} \\

Qwen-Image~\cite{wu2025qwenimagetechnicalreport}
& \bluecellnp{70}{0.99} & 0.92 & \bluecellnp{100}{0.89} & 0.88 & \bluecellnp{70}{0.76} & \bluecellnp{100}{0.77} & \bluecellnp{70}{0.87}
& 52.02 & 74.44 & 34.13 & 53.53 \\

\bottomrule[1.5pt]
\end{tabular}

\end{threeparttable}}
\label{tab:geneval_unipercept_full}
\end{table*}

\noindent\textbf{For Datasets.}
We evaluate a variety of natural-image and AIGC-image datasets using UniPercept as the assessment metric, with results summarized in Tab.\ref{tab:metrics_dataset}. Among all datasets, Unsplash~\cite{unsplash} and Blip3o-60K\cite{chen2025blip3ofamilyfullyopen} achieve the strongest overall performance across the three domains of IAA, IQA, and ISTA. A more in-depth investigation of these distributional characteristics is left for future work.

\begin{table}[htbp]
\centering
\small
\caption{\textbf{UniPercept Metrics on Various Datasets.}}
\vspace{-6pt}
\label{tab:metrics_dataset}
\resizebox{0.49\textwidth}{!}{
\begin{threeparttable}[t]
\renewcommand{\arraystretch}{0.9}

\begin{tabular}{l|cccc}
\toprule[1.5pt]
\textbf{Dataset} & \textbf{UniPercept-IAA} & \textbf{UniPercept-IQA} & \textbf{UniPercept-ISTA} & \textbf{Avg.} \\
\midrule

\multicolumn{5}{l}{\textit{\textbf{Natural Images}}} \\
ImageNet~\cite{russakovsky2015imagenetlargescalevisual}   
& 53.88 & 61.90 & 36.79 & 50.85 \\

Unsplash~\cite{unsplash}   
& \bluecellnp{70}{62.49} & 69.19 & \bluecellnp{70}{43.32} & \bluecellnp{70}{58.33} \\

DF2K~\cite{df2k1,df2k2,df2k3,df2k4}       
& 45.99 & 52.92 & 34.78 & 44.56 \\

LAION-5B~\cite{laion5b}   
& 60.56 & \bluecellnp{70}{69.21} & 38.85 & 56.21 \\

\midrule
\multicolumn{5}{l}{\textit{\textbf{AIGC Images}}} \\
Blip3o-60K~\cite{chen2025blip3ofamilyfullyopen} 
& \bluecellnp{100}{63.81} & \bluecellnp{100}{73.88} & \bluecellnp{100}{49.38} & \bluecellnp{100}{62.36} \\

ImgEdit~\cite{ye2025imgeditunifiedimageediting}    
& 55.83 & 59.77 & 36.88 & 50.83 \\

\bottomrule[1.5pt]
\end{tabular}

\end{threeparttable}}
\vspace{-15pt}
\end{table}

\section{Conclusion}
\label{sec:Conclusion}

We provide \textbf{UniPercept-Bench}, a unified benchmark built on a hierarchical definition for perceptual-level image understanding. 
We also develop a strong baseline \textbf{UniPercept} through Domain-Adaptive Pre-training and Task-Aligned RL, which generalizes well across perceptual domains and outperforms existing MLLMs. 
UniPercept further serves as a plug-and-play \textbf{reward model} for perceptually aligned post-training of T2I models, enabling controllable improvements in perceptual attributes. Beyond model optimization, it also offers a unified perceptual diagnostic tool that reveals systematic behavioral and dataset-level patterns, highlighting its broader utility for future perceptual-level research.

\noindent\textbf{Limitations.}
Although UniPercept-Bench is sufficiently large for current perceptual-level tasks, it remains smaller than typical semantic-level benchmarks. Further expansion in scale will be explored in future work.   

{
    \small

}

\maketitlesupplementary

\section{Unifying Perceptual-Level Image Understanding}

\subsection{From Semantics to Perception}
High-level image understanding concerns \emph{what} an image depicts—objects, actions, scenes, and their semantic relations (e.g., ``a dog running on grass,'' ``a busy street at night''). 
In contrast, \textbf{perceptual-level image understanding} concerns \emph{how} an image looks and feels to human observers. 
It captures intrinsic perceptual properties of the visual signal, largely independent of whether semantic content is correctly recognized.

An image may be semantically understandable yet perceptually flawed due to poor composition, compression artifacts, or unrealistic textures. 
Perceptual-level understanding emphasizes visual cues and their human-perceived effects—beauty and preference, fidelity and distortion, and structural/textural realism—rather than categorical recognition or reasoning. 
This distinction is well aligned with prior research separating perceptual judgments from semantic interpretation.

We view perceptual-level image understanding as a unified family of human-aligned, low-level perceptual tasks, consisting of:
\begin{itemize}
    \item \textbf{Image Aesthetics Assessment (IAA)}
    \item \textbf{Image Quality Assessment (IQA)}
    \item \textbf{Image Structure \& Texture Assessment (ISTA)}
\end{itemize}


Together, these tasks form a top-down to bottom-up characterization of perceptual attributes:
\begin{itemize}
\item \textbf{Aesthetics} captures holistic human preference and artistic visual appeal, reflecting composition, mood, and overall expressive quality.
\item \textbf{Quality} captures visual fidelity and perceived degradation, encompassing distortions and signal-level clarity.
\item \textbf{Structure \& Texture} capture local cues shaping realism and richness, including fine-grained patterns, materials, and small-scale structural details.
\end{itemize}

This tripartite formulation provides a more complete perceptual description than treating aesthetics, quality, or texture as isolated benchmarks.


\subsection{Detailed Explanation of IAA, IQA, and ISTA}

\subsubsection{Image Aesthetics Assessment (IAA)}
IAA represents the highest level of perceptual abstraction. 
Its goal is to predict an image’s holistic aesthetic appeal---how pleasing, expressive, or artistically valuable it appears. 
The task draws on both philosophical aesthetics and empirical studies of human visual preference~\cite{wiki_iaa}.

Common aesthetic factors include:~\cite{artimuse}
\begin{itemize}
    \item \textbf{Composition:} balance, framing, symmetry, rule-of-thirds, leading lines;
    \item \textbf{Visual Elements:} harmony, exposure, atmosphere;
    \item \textbf{Creativity:} intent, novelty, genre coherence;
    \item \textbf{Emotion:} theme communication, storytelling.
\end{itemize}

Because aesthetic judgments depend strongly on viewer psychology, IAA is the most subjective perceptual dimension. 

\subsubsection{Image Quality Assessment (IQA)}
IQA bridges subjective perception and objective image formation. 
Its goal is to estimate perceived visual fidelity, i.e., the severity of degradations relative to an ideal image. 
Such degradations commonly arise from sensor noise, blur, compression, transmission errors, and artifacts~\cite{wiki_iqa}.

Whereas IAA asks ``Is this beautiful?'', IQA asks ``Is this technically clean?'' Partially, IQA evaluates:~\cite{depictqa_v1,depictqa_v2,Q-bench,q-ground}
\begin{itemize}
    \item \textbf{Signal Fidelity:} sharpness, exposure, color naturalness;
    \item \textbf{Artifacts:} blur, noise, compression artifacts, aliasing;
    \item \textbf{Perceptual Faithfulness:} how ``undistorted'' the image appears.
\end{itemize}

These properties have both measurable signal correlates and human-dependent thresholds, placing IQA between IAA and ISTA in subjectivity.


\subsubsection{Image Structure \& Texture Assessment (ISTA)}
ISTA measures local perceptual primitives that determine whether an image appears structurally coherent and texturally rich~\cite{wiki_ista}. 
Unlike the global emphasis of IAA and IQA, ISTA focuses on pixel- and patch-level cues, including:
\begin{itemize}
    \item \textbf{Local Structure:} edges, contours, geometric consistency;
    \item \textbf{Texture Statistics:} granularity, repetitiveness, roughness/smoothness;
    \item \textbf{Material-like Cues:} surface micro-patterns conveying realism (e.g., fabric weave, wood grain).
\end{itemize}

ISTA connects to classic studies of texture perception and structural similarity, but reframes them as a unified perceptual assessment dimension that many vision models underrepresent.


\subsection{The Perceptual Hierarchy}
IAA, IQA, and ISTA form a complementary perceptual hierarchy:
\begin{itemize}
    \item \textbf{IAA:} holistic beauty and preference (highly subjective),
    \item \textbf{IQA:} technical fidelity and distortion\\ (mixed objective/subjective),
    \item \textbf{ISTA:} fine-grained structure and texture realism\\ (more objective).
\end{itemize}

These dimensions are complementary but non-redundant. High quality does not imply high aesthetics; rich texture does not imply high quality; high aesthetics does not imply rich texture.

Thus, no single dimension adequately explains human perceptual judgments.


\subsection{Why Unify Perceptual Understanding?}
Unifying IAA, IQA, and ISTA provides key scientific and practical benefits:

\begin{itemize}
    \item \textbf{Shared Perceptual Representations:} joint learning encourages a universal perceptual embedding.
    \item \textbf{Better Evaluators for Generative/Restoration Models:} unified perceptual signals diagnose model failures beyond semantic metrics.
    \item \textbf{Controllable Generation/Editing:} separate perceptual axes support targeted improvements (e.g., enhancing texture without harming quality).
    \item \textbf{Dataset Curation:} perceptual scoring allows filtering or reweighting data by aesthetics, fidelity, or structure.
    \item \textbf{Human-centric Applications:} real-world systems optimize perceived quality, not just semantics.
\end{itemize}


\subsection{Our Contributions}
We introduce a unified paradigm for perceptual-level image understanding. Our contributions include:
\begin{itemize}
    \item \textbf{Unified Task Definition:} a coherent taxonomy integrating IAA, IQA, and ISTA. More details are provided in Tab.~\ref{tab:definition_iaa}, Tab.~\ref{tab:definition_iqa}, Tab.~\ref{tab:definition_ista} and Sec.~\ref{sec:definitin_system}.
    \item \textbf{Holistic Evaluation Protocol:} performance measurement across three perceptual dimensions.
    \item \textbf{Unified Baseline Model:} a multi-task perceptual model predicting aesthetics, quality, and structure/texture jointly.
    \item \textbf{Downstream Validation:} unified perceptual signals improve applications such as reward modeling for post-training.
\end{itemize}

\section{Details of ISTA}
\label{sec:ista_detail}

\subsection{Structural Annotation}
\label{sec:structure_ista}

We establish a systematic definition of image structure and texture, as presented in Tab.~\ref{tab:definition_ista}. Building upon this definition framework, we design Structural Annotation scheme for ISTA as shown in Fig.~\ref{fig:structural_annotation}. We also employ the following prompt to guide the MLLM in performing the Structural Annotation for ISTA.

\begin{figure}[htbp]
\centering
\footnotesize
\begin{lstlisting}[style=jsonstyle]
{
    "SceneType": "<SceneType>",
    "SceneName": "<SceneName>",
    "Components": [
        {
            "ComponentName": "<Component_1>",
            "DescriptionContent": {
                "PhysicalStructure": {
                    "BaseMorphology": ["<Morphology_1>"],
                    "Arrangement": ["<Arrangement_1>"]
                },
                "MaterialRepresentation": {
                    "MaterialClass": ["<MaterialClass_1>"],
                    "SurfaceProperties": ["<SurfaceProperty_1>"]
                },
                "GeometricComposition": {
                    "PlanarContour": ["<PlanarContour_1>"],
                    "VolumetricForm": ["<VolumetricForm_1>"]
                },
                "SemanticPerception": {
                    "FunctionalInference": ["<FunctionalInference_1>"],
                    "StyleType": ["<StyleType_1>"]
                }
            }
        },
        {
            "ComponentName": "<Component_2>",
                ...
            }
        }
    ]
}

\end{lstlisting}
\caption{Template of ISTA Structural Annotation.}
\label{fig:structural_annotation}
\end{figure}


\begin{promptbox}[Prompt for ISTA Structural Annotation]
   [PRIOR KNOWLEDGE BASE]

    - Base Morphology  
    
    blotchy, braided, bubbly, bumpy, chequered, cobwebbed, cracked, crosshatched, crystalline, dotted, fibrous, flecked, freckled, frilly, grid, grooved, honeycombed, interlaced, knitted, lacelike, lined, marbled, matted, meshed, paisley, perforated, pitted, pleated, porous, scaly, smeared, spiralled, sprinkled, stratified, striped, studded, swirly, veined, woven, wrinkled, zigzagged, smooth

    - Material Type  
    
    1. Natural Materials:  
        Foliage, Grass, Skin, Stone, Wood, Water, Hair 
        
    2. Man-Made Materials:  
        Brick, Carpet, Ceramic, Fabric, Glass, Leather, Metal, Mirror, Painted Surface, Paper, Plastic, Polished Stone, Tile, Wallpaper, Concrete, Food Surface  
        
    3. Environmental / Background Textures:  
        Sky, Clouds, Fog / Mist  

    - Two-Dimensional Shape  
    
    Rectangle, Square, Circle, Ellipse / Oval, Triangle, Equilateral Triangle, Isosceles Triangle, Scalene Triangle, Right Triangle, Trapezoid / Trapezium, Parallelogram, Rhombus, Pentagon, Hexagon, Heptagon, Octagon, Nonagon, Decagon, Star, Pentagram, Hexagram, Cross, Arrow, Semicircle, Sector, Crescent, Annulus / Ring, Heart, Lemniscate, Lune / Bow Shape, Spiral, Waveform, Teardrop  

    - Three-Dimensional Shape Categories  
    
    Sphere, Ellipsoid, Cube, Cuboid, Cylinder, Cone, Pyramid, Tetrahedron, Octahedron, Dodecahedron, Icosahedron, Prism, Triangular Prism, Rectangular Prism, Pentagonal Prism, Hexagonal Prism, Torus, Annular Torus, Paraboloid, Hyperboloid, Elliptic Cylinder, Hyperbolic Cylinder, Truncated Cone, Truncated Pyramid, Capsule, Dome, Lens, Bipyramid, Frustum, Möbius Strip, Knot, Klein Bottle  

    - Style Semantics  
    
    Embossed, Engraved, Rough, Smooth, Matte, Glossy, Brushed, Honeycomb, Geometric, Fractal, Tile Mosaic, Chinese Cloud Pattern, Dragon Scale, Cyberpunk Holographic, Steampunk Mechanical  

    [STRUCTURE TEMPLATE]  
    
    - Scene Decomposition Principles  
    
    A. Single Scene: (Please introduce the description object.)  
    
    B. Composite Scene: (Please introduce the different objects. Describe the object separately)  
    
    (e.g., street → architectural cluster + pavement system + sky background)  

    - Description Content (In Composite Scene mode, Please indicate the object in a single description.)  
    
    (e.g., Description Content [street])  

    1. Physical Structure  
    
        Base Morphology (*) → Select 1-3 terms from the lexicon (Basically comes from Base Morphology) to describe the surface texture (Focus on texture form rather than shape)
        Arrangement (!) → Describe spatial layout or directionality of texture: orientation (e.g., diagonal, radial), distribution pattern (e.g., clustered, layered), or density changes (optional)  
        
        Dynamics (!) → Motion/transition states (optional)  

    2. Material Representation  
    
        Material Class (*) → Select from Material Type 
        
        Surface Properties (!) → Reflectivity/translucency (optional)  

    3. Geometric Composition  
    
        Planar Contour (!) → 2D shape terms from Two-Dimensional Shape (where applicable)  (optional)  
        
        Volumetric Form (!) → 3D form terms from Three-Dimensional Shape Categories (where applicable)  (optional)  

    4. Semantic Perception  
    
        Functional Inference (!) → Only!!! with text/icons present (e.g. The icon is: no traffic, and the text is: XX Street...)  (optional)  
        
        Style Type (!) → Must use style semantics terms (e.g. Baroque style, Xiangyun style...) (optional)

    [Execution Standards]  
    
    1. Terminology Enforcement: (*)-marked fields require exact term matches
    
    2. Format Purity: Output only structured content, no explanations  
    
    3. Hierarchy Preservation: Apply complete template per independent unit  
    
    4. Complexity Adaptation: Single description for simple objects, multi-unit decomposition for complex scenes  
    
    5. Lexicon Flexibility: For *-marked fields, use official lexicon terms where possible. 
    Free-form extensions are allowed if they add necessary clarity or express phenomena outside the vocabulary.  
    
    6. Mixed Mode Expression: Structured descriptions may combine fixed taxonomy terms with precise natural language when facing edge cases or fine-grained observations.  

    \begin{lstlisting}[style=jsonstyle]
[Example]
{
    "SceneType": "Composite Scene",
    "SceneName": "Urban skyscraper cluster",
    "Components": [
        {
            "ComponentName": "Buildings",
            "DescriptionContent": {
                "PhysicalStructure": {
                    "BaseMorphology": ["grid"],
                    "Arrangement": ["Vertical,Symmetrical"],
                    "Dynamics": ["N/A"]
                },
                "MaterialRepresentation": {
                    "MaterialClass": ["Glass","Metal"],
                    "SurfaceProperties": ["Glossy","Reflective"]
                },
                "GeometricComposition": {
                    "PlanarContour": ["Rectangle"],
                    "VolumetricForm": ["Cuboid"]
                },
                "SemanticPerception": {
                    "FunctionalInference": ["N/A"],
                    "StyleType": ["Modern Architecture"]
                }
            }
        },
        {
            "ComponentName": "Sky Background",
            "DescriptionContent": {
                "PhysicalStructure": {
                    "BaseMorphology": ["smooth"],
                    "Arrangement": ["N/A"],
                    "Dynamics": ["N/A"]
                },
                "MaterialRepresentation": {
                    "MaterialClass": ["Sky"],
                    "SurfaceProperties": ["N/A"]
                },
                "GeometricComposition": {
                    "PlanarContour": ["N/A"],
                    "VolumetricForm": ["N/A"]
                },
                "SemanticPerception": {
                    "FunctionalInference": ["N/A"],
                    "StyleType": ["N/A"]
                }
            }
        }
    ]
}
\end{lstlisting}
    
    Please use the above foundational knowledge and the provided example to perform a texture and structural analysis of the image.   
\end{promptbox}

\subsection{Definition of ISTA-10K}
\label{sec:definition_ista}
ISTA-10K is our curated visual rating dataset for evaluating the structure–texture dimension of images.
The ISTA score quantifies an image’s \textit{structure–texture richness}, reflecting both
(1) the \textbf{complexity and diversity of its visual textures}, and
(2) the \textbf{richness and organization of its structural components}.
Images exhibiting more varied textures, materials, geometric forms, and semantically distinct elements are assigned higher ISTA scores.

\noindent\textbf{Texture Intensity Mapping.}
To quantify the complexity of base morphological patterns, each texture term $t$ is assigned 
a discrete weight reflecting its perceived structural richness. 
Lower weights correspond to visually simple and uniform textures, whereas higher weights 
indicate more irregular, high-frequency, or structurally complex surface patterns.  
The weighting function is defined as:
\begin{equation}
w(t)=
\begin{cases}
1, & t\in\mathcal{T}_{\text{weak}},\\
2, & t\in\mathcal{T}_{\text{medium}},\\
3, & t\in\mathcal{T}_{\text{strong}},\\
0, & \text{otherwise}.
\end{cases}
\label{eq:texture_weight}
\end{equation}

\noindent The three texture sets used in Eq.~\ref{eq:texture_weight} are summarized in 
Table~\ref{tab:texture_sets}. 
These categories are derived from commonly observed visual morphologies and are grouped by 
increasing structural complexity. Weak textures are visually simple and low-variation patterns with minimal geometric or material irregularities.
Medium textures exhibit moderate complexity with richer local variations and more structured arrangements.
Strong textures reflect high morphological complexity, featuring irregular, multi-scale, or highly distinctive patterns often associated with heterogeneous or fine-grained natural structures.

\begin{table}[h]
\centering
\small
\caption{Texture categories used for texture intensity mapping. Higher groups indicate richer and more complex morphology patterns.}
\vspace{-6pt}
\label{tab:texture_sets}
\begin{tabular}{p{2.0cm} p{5.0cm}}
\toprule
\textbf{Category (Weight)} & \textbf{Texture Terms} \\
\midrule
\textbf{Weak (1)} &
smooth, plain, uniform, lined, grid, striped, chequered, dotted, freckled \\
\midrule
\textbf{Medium (2)} &
braided, woven, crosshatched, meshed, cobwebbed, lacelike, knitted, spiralled, swirly \\
\midrule
\textbf{Strong (3)} &
bumpy, blotchy, bubbly, cracked, crystalline, flecked, frilly, grooved, honeycombed, 
marbled, matted, paisley, perforated, pitted, pleated, porous, scaly, smeared, 
sprinkled, stratified, studded, veined, wrinkled, zigzagged \\
\bottomrule
\end{tabular}
\end{table}

\noindent\textbf{Component-Level ISTA Rating.}
For each component $c$, we compute:
\begin{equation}
S(c)=S_{\text{PS}}(c)+S_{\text{MR}}(c)+S_{\text{GC}}(c)+S_{\text{SP}}(c).
\end{equation}

These terms quantify different dimensions of structural and perceptual richness:
\begin{itemize}
    \item \textbf{Physical Structure (PS):} describes the surface morphology and spatial arrangement of structural patterns;
    \item \textbf{Material Representation (MR):} captures the diversity of material classes and their surface properties;
    \item \textbf{Geometric Composition (GC):} reflects variations in planar contours and volumetric forms;
    \item \textbf{Semantic Perception (SP):} accounts for functional cues and stylistic categories associated with the component
\end{itemize}
Each sub-score is defined as follows (all ``N/A'' entries excluded):

\begin{align}
S_{\text{PS}}(c)
&=\sum_{t\in \text{BaseMorphology}(c)} w(t)
  +\bigl|\text{Arrangement}(c)\bigr|,\\[3pt]
S_{\text{MR}}(c)
&=\bigl|\text{MaterialClass}(c)\bigr|
  +\bigl|\text{SurfaceProperties}(c)\bigr|,\\[3pt]
S_{\text{GC}}(c)
&=\bigl|\text{PlanarContour}(c)\bigr|
  +\bigl|\text{VolumetricForm}(c)\bigr|,\\[3pt]
S_{\text{SP}}(c)
&=\bigl|\text{FunctionalInference}(c)\bigr|
  +\bigl|\text{StyleType}(c)\bigr|.
\end{align}

\vspace{4pt}
\noindent\textbf{Image-Level ISTA Rating.}
If the image contains a set of components $\mathcal{C}$,
\begin{equation}
S_{\text{ISTA}}
=|\mathcal{C}| + \sum_{c\in\mathcal{C}} S(c).
\end{equation}

For images without explicit components, the whole image is treated as a single component:
\begin{equation}
S_{\text{ISTA}} = 1 + S(c_{\text{image}}).
\end{equation}

Finally, the score is clipped to match the 0--100 rating range:
\begin{equation}
S_{\text{ISTA}} \leftarrow \min(S_{\text{ISTA}},\,100).
\end{equation}

This formulation yields a deterministic and interpretable measure of structure--texture richness that aligns with the hierarchical annotation schema in UniPercept-Bench.

\section{Details of UniPercept-Bench}
\label{sec:benchmark_detail}

\subsection{Definition System}
\label{sec:definitin_system}
The UniPercept definition system is organized into three levels: \textit{Domain–Category–Criterion}. The complete set of fine-grained attributes and their brief descriptions are provided in Tab.~\ref{tab:definition_iaa}, Tab.~\ref{tab:definition_iqa} and Tab.~\ref{tab:definition_ista}.

\begin{table*}[t]
\centering
\small
\caption{\textbf{Definition details of IAA domain.}}
\vspace{-6pt}
\renewcommand{\arraystretch}{1.10}

\resizebox{0.93\textwidth}{!}{
\begin{tabular}{p{0.8cm} p{4cm} p{4.5cm} p{8.0cm}}
\toprule[1.5pt]
\textbf{No.} & \textbf{Category} & \textbf{Criterion} & \textbf{Description} \\
\midrule
1 & Composition \& Design (\textbf{\textit{Comp.}}) & Visual Balance & Assess the distribution of visual elements such as shapes, tones, and colors to determine equilibrium across the frame. \\
2 & Composition \& Design & Hierarchical Emphasis & Evaluate the relative prominence of visual elements based on size, contrast, or positioning. \\
3 & Composition \& Design & Structural Organization & Examine spatial alignment, grid conformity, and grouping of elements within the image frame. \\
4 & Composition \& Design & Compositional Rhythm & Assess repetition, spacing, and directional continuity of elements that suggest visual pacing. \\
5 & Composition \& Design & Harmonic Unity & Evaluate visual consistency in shape proportions, relative sizes, and orientation patterns. \\
6 & Composition \& Design & Composition \& Design Level & Indicates the overall compositional quality — reflecting balance, rhythm, and structural harmony. \\
\midrule
7 & Visual Elements \& Structure (\textbf{\textit{VisStr.}})  & Line Dynamics & Assess structure, orientation, and density of linework contributing to visual form. \\
8 & Visual Elements \& Structure & Shape Clarity & Assess clarity of 2D shape boundaries and separation from background. \\
9 & Visual Elements \& Structure & Form Realization & Evaluate rendering of 3D form through shading, lighting gradients, and perspective cues. \\
10 & Visual Elements \& Structure & Spatial Illusion & Judge depth cues such as occlusion, scale variation, and linear perspective. \\
11 & Visual Elements \& Structure & Light Modeling & Assess consistency and realism of lighting, highlights, and shadows. \\
12 & Visual Elements \& Structure & Visual Elements \& Structure Level & Measures mastery of fundamental visual elements including lines, shapes, form, and spatial coherence. \\
\midrule
13 & Technical Execution (\textbf{\textit{Tech.}}) & Material Proficiency & Evaluate control and precision in using the medium, including brushstroke discipline. \\
14 & Technical Execution & Rendering Precision & Assess refinement in edge definition, gradient transitions, and micro-level detailing. \\
15 & Technical Execution & Focus Control & Judge sharpness, blur, or depth-of-field usage to structure visual hierarchy. \\
16 & Technical Execution & Tonal and Exposure Control & Evaluate luminance distribution and tonal range. \\
17 & Technical Execution & Technical Execution Level & Represents technical proficiency — tonal control, precision, and rendering quality. \\
\midrule
18 & Originality \& Creativity (\textbf{\textit{Creat.}}) & Concept Innovation & Assess distinctiveness of concept or narrative. \\
19 & Originality \& Creativity & Creative Problem-Solving & Evaluate ingenuity in visual execution or compositional decision-making. \\
20 & Originality \& Creativity & Originality \& Creativity Level & Expresses creative strength — innovation, imagination, and conceptual uniqueness. \\
\midrule
21 & Theme \& Communication (\textbf{\textit{Theme.}}) & Subject Clarity & Evaluate clarity of subject or message. \\
22 & Theme \& Communication & Narrative Depth & Assess symbolic or narrative layering. \\
23 & Theme \& Communication & Cultural Insight & Judge engagement with cultural, historical, or social ideas. \\
24 & Theme \& Communication & Theme \& Communication Level & Evaluates thematic clarity and communication effectiveness. \\
\midrule
25 & Emotion \& Viewer Response (\textbf{\textit{Emo.}}) & Emotional Resonance & Evaluate emotional tone and viewer response. \\
26 & Emotion \& Viewer Response & Viewer Engagement & Assess long-term viewer compellingness. \\
27 & Emotion \& Viewer Response & Interpretive Openness & Judge clarity–ambiguity balance inviting interpretation. \\
28 & Emotion \& Viewer Response & Emotion \& Viewer Response Level & Measures emotional and psychological impact. \\
\midrule
29 & Overall Gestalt (\textbf{\textit{Gest.}}) & Holistic Cohesion & Evaluate integration of visual and conceptual components into a unified whole. \\
30 & Overall Gestalt & Overall Gestalt Level & Represents holistic integration across all components. \\
\midrule
31 & Comprehensive Evaluation (\textbf{\textit{CompEv.}}) & Comprehensive Evaluation Level & Synthesizes all artistic dimensions — perceptual quality and conceptual depth. \\
\bottomrule[1.5pt]

\end{tabular}
} 
\label{tab:definition_iaa}
\end{table*}

\begin{table*}[t]
\centering
\small
\caption{\textbf{Definition details of IQA domain.}}
\vspace{-6pt}
\renewcommand{\arraystretch}{1.10}

\resizebox{0.95\textwidth}{!}{
\begin{tabular}{p{0.8cm} p{4cm} p{4.5cm} p{8.0cm}}
\toprule[1.5pt]
\textbf{No.} & \textbf{Category} & \textbf{Criterion} & \textbf{Description} \\
\midrule
1 & Distortion Location (\textbf{\textit{Loc.}}) & Location Description &
Precisely identify and describe specific spatial regions within the image where distortions are visible or concentrated. The question must explicitly reference a concrete part of the image, not the image as a whole. \\
2 & Distortion Location & Object Association &
Specify the semantic or structural objects in the image that are impacted or altered by the distortions. The question must explicitly reference a concrete object, not just a general distortion type. \\
\midrule
3 & Distortion Severity (\textbf{\textit{Sev.}}) & Severity Level &
Evaluate the severity of distortion in the image as None, Slight, or Obvious, reflecting visibility and perceptual impact. \\
\midrule
4 & Distortion Type (\textbf{\textit{Type.}}) & Distortion Types Present &
Identify distortion types present in the scene from a comprehensive taxonomy, including blur, noise, compression, brightness, contrast, saturation, sharpening, quantization, exposure, pixelation, color diffusion, jitter, transmission errors, and multi-distortion combinations. \\
\bottomrule[1.5pt]
\end{tabular}
} 
\label{tab:definition_iqa}
\end{table*}

\begin{table*}[t]
\centering
\small
\caption{\textbf{Definition details of ISTA domain.}}
\vspace{-6pt}
\renewcommand{\arraystretch}{1.10}

\resizebox{0.95\textwidth}{!}{
\begin{tabular}{p{0.8cm} p{4cm} p{4.5cm} p{8.0cm}}
\toprule[1.5pt]
\textbf{No.} & \textbf{Category} & \textbf{Criterion} & \textbf{Description} \\
\midrule
1 & Scene Decomposition Principles (\textbf{\textit{Scene.}}) & Scene Classification &
Classify the scene as (A) Single-Object Scene with one primary subject, or (B) Composite Scene containing multiple distinguishable components. Describe the main object(s) clearly. \\
\midrule
2 & Physical Structure (\textbf{\textit{Phys.}}) & Base Morphology &
Describe surface texture using perceptual descriptors such as fibrous, grooved, marbled, veined, smooth, etc. \\
3 & Physical Structure & Spatial Arrangement &
Describe texture orientation, distribution pattern, and density variation across regions (e.g., horizontal, clustered, layered, radial, uniform). \\
\midrule
4 & Material Representation (\textbf{\textit{Mat.}}) & Material Identification &
Identify the perceived material category using a standardized taxonomy (natural, man-made, or environmental materials) present in the image. \\
5 & Material Representation & Surface Behavior &
Describe optical surface properties such as glossiness, translucency, or matte finish. \\
\midrule
6 & Geometric Composition (\textbf{\textit{Geo.}}) & 2D Contour &
Classify the 2D outline shape using a standardized lexicon including basic shapes, polygons, special forms, or organic/curved forms. \\
7 & Geometric Composition & 3D Volume &
Describe the implied 3D volumetric form using a taxonomy of basic solids, polyhedra, prisms, or complex mathematical shapes. \\
\midrule
8 & Semantic Perception (\textbf{\textit{Sem.}}) & Functional Suggestion &
Infer functional or symbolic implications of textures/motifs based on appearance, referencing standardized functional texture/style descriptors. \\
9 & Semantic Perception & Stylistic Classification &
Assign a stylistic category (e.g., Minimalist, Gothic, Art Deco, Futuristic, Cyberpunk, Chinese Cloud Pattern) based on visual elements and decorative cues. \\
\bottomrule[1.5pt]
\end{tabular}
} 
\label{tab:definition_ista}
\vspace{-10pt}
\end{table*}

\begin{table*}[!t]
\centering
\small
\caption{\textbf{Comparison with existing benchmarks \& datasets.}}
\vspace{-6pt}
\resizebox{2\columnwidth}{!}{
\begin{threeparttable}
\renewcommand{\arraystretch}{1.05}
\begin{tabular}{l|c|c|c|c|c|ccc|cc}
\toprule[1.2pt]
\multirow{2}{*}{\textbf{Benchmark}} & \multirow{2}{*}{\textbf{\# Test}} & \multirow{2}{*}{\textbf{\# QA Category}} & \multirow{2}{*}{\textbf{Data Format}} & \multirow{2}{*}{\textbf{Annotation Level}} & \multirow{2}{*}{\textbf{Annotator}} & \multicolumn{3}{c|}{\textbf{Perceptual-Level Domain}} & \multicolumn{2}{c}{\textbf{Task}} \\
\cmidrule(lr){7-9} \cmidrule(lr){10-11}
& & & & & & \textit{IAA} & \textit{IQA} & \textit{ISTA} & \textbf{VQA} & \textbf{VR} \\
\midrule
\multicolumn{2}{l}{\textit{\textbf{VQA Benchmarks \& Datasets}}} \\
Q-Bench~\cite{Q-bench} & $\sim$1.5K  & -- & Text & Category-Level & Human & -- & \checkmark & -- & \checkmark & -- \\
AesBench~\cite{Aesbench} & $\sim$10K & 10 & Text & Example-Level & Human & \checkmark & -- & -- & \checkmark & -- \\
DQ-495K~\cite{depictqa_v2} & $\sim$56K & -- & Text & Category-Level & Human \& MLLM & -- & \checkmark & -- & \checkmark & -- \\
Q-Instruct-DB~\cite{q-instuct} & -- & -- & Text & Category-Level & Human \& MLLM & -- & \checkmark & -- & \checkmark & -- \\
Co-Instruct~\cite{co-instuct} & -- & -- & Text & Category-Level & Human \& MLLM & -- & \checkmark & -- & \checkmark & -- \\
Q-Ground-100K~\cite{q-ground} & $\sim$1K & -- & Text & Category-Level & Human \& MLLM & -- & \checkmark & -- & \checkmark & -- \\
\midrule
\multicolumn{2}{l}{\textit{\textbf{VR Benchmarks \& Datasets}}} \\
ArtiMuse-10K~\cite{artimuse} & $\sim$1K & 15 & Rating \& Text & Example-Level & Human & \checkmark & -- & -- & -- & \checkmark \\
AVA~\cite{ava} & $\sim$20K & -- & Rating & Example-Level & Human & \checkmark & -- & -- & -- & \checkmark \\
TAD66K~\cite{tad66k} & $\sim$15K & -- & Rating & Example-Level & Human & \checkmark & -- & -- & -- & \checkmark \\
KonIQ-10K~\cite{koniq} & $\sim$2K & -- & Rating & Example-Level & Human & -- & \checkmark & -- & -- & \checkmark \\
SPAQ~\cite{spaq} & $\sim$1K & -- & Rating & Example-Level & Human & -- & \checkmark & -- & -- & \checkmark \\
KADID~\cite{kadid10k} & $\sim$1K & -- & Rating & Example-Level & Human & -- & \checkmark & -- & -- & \checkmark \\
\midrule
\textbf{UniPercept-Bench (Ours)} & \textbf{$\sim$6K} & \textbf{44} & \textbf{Rating \& Text} & \textbf{Example-Level} & \textbf{Human \& MLLM} & \textbf{\checkmark} & \textbf{\checkmark} & \textbf{\checkmark} & \textbf{\checkmark} & \textbf{\checkmark} \\
\bottomrule[1.2pt]
\end{tabular}
\end{threeparttable}}
\vspace{-6pt}
\label{tab:comparison_with_othe_benchmarks}
\end{table*}

\subsection{Evaluation Details}
\label{sec:evaluation_detail}

For the VQA task, we directly feed the image, question, and answer options into the MLLM during evaluation.
For the VR task, if a model provides dedicated interfaces for IAA or IQA scoring, we invoke them directly.
For models without such interfaces, we evaluate them using the following prompts:

\begin{promptbox}[Prompt for Visual Rating (IAA)]
Please rate the aesthetics of this image and provide a score between 0 and 100, where 0 represents the lowest quality and 100 represents the highest. Your response should contain only an integer value.
\end{promptbox}

\begin{promptbox}[Prompt for Visual Rating (IQA)]
Please rate the quality of this image and provide a score between 0 and 100, where 0 represents the lowest quality and 100 represents the highest. Your response should contain only an integer value.
\end{promptbox}

\begin{promptbox}[Prompt for Visual Rating (ISTA)]
Please rate the structure and texture richness of this image and provide a score between 0 and 100, where 0 represents the lowest quality and 100 represents the highest. Your response should contain only an integer value.
\end{promptbox}

\subsection{Comparison with Other Benchmarks}
\label{sec:comparions_benchmark}

As shown in Tab.~\ref{tab:comparison_with_othe_benchmarks}, we compare UniPercept-Bench with other widely-used benchmarks in image assessment. UniPercept-Bench provides finer-grained QA categories, supports both rating and textual formats, and employs a scalable annotation pipeline that combines human expertise with MLLM assistance to achieve example-level annotation. It further covers all three perceptual-level domains and includes both VQA and VR tasks, making it the most comprehensive and well-defined benchmark for perceptual-level image understanding to date.


\subsection{Relations Among Different Perceptual Domains}
\label{sec:domain_relation}
\begin{figure}[htbp]
    \centering
    \includegraphics[width=0.49\textwidth]{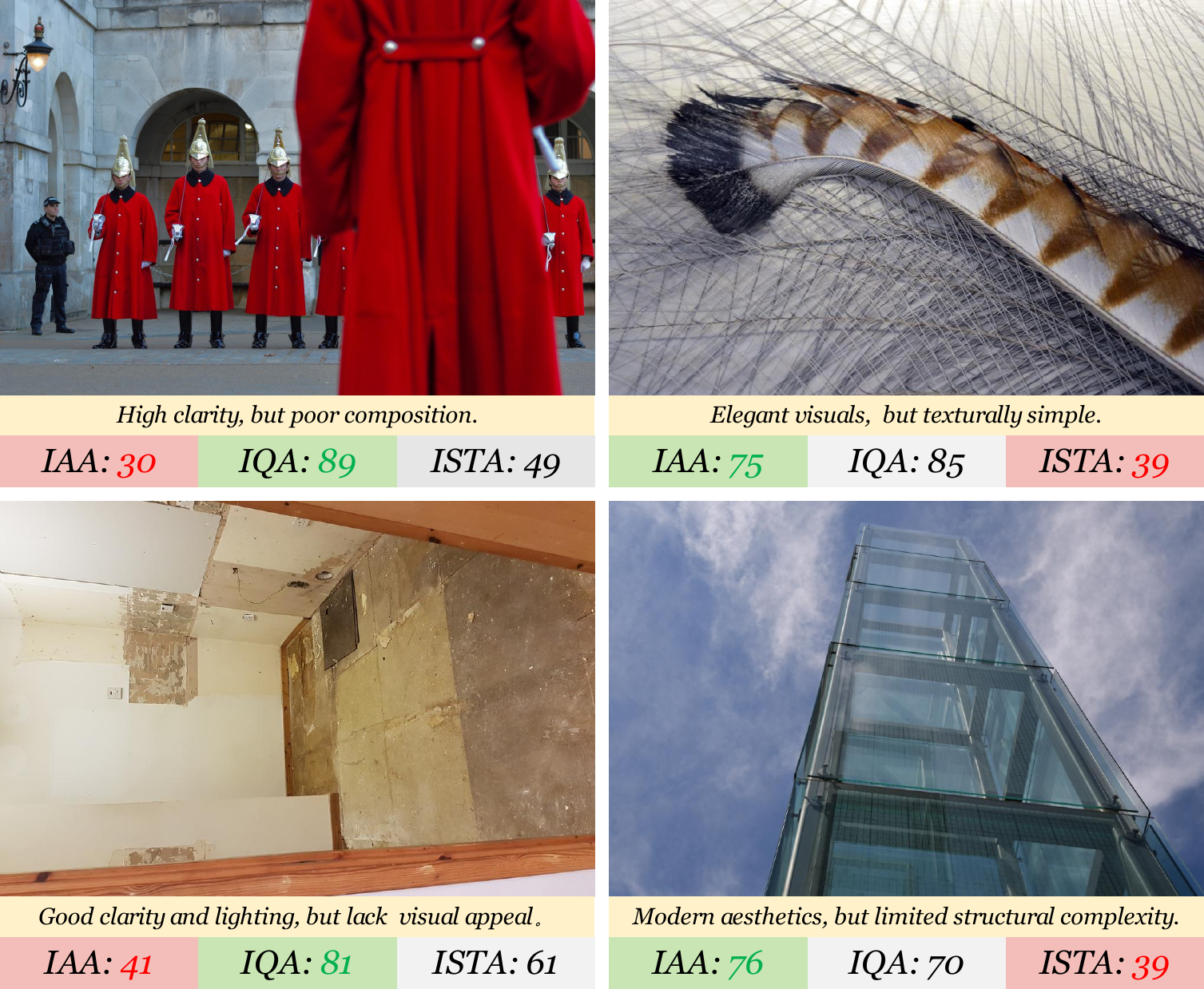}
    \vspace{-10pt}
    \caption{\textbf{Relationship between different perceptual domains}. Ratings are provided by UniPercept.}
    \label{fig:relationship}
    \vspace{-20pt}
\end{figure}

As discussed in Sec.3.1 of the main paper, the three domains (IAA, IQA, and ISTA) characterize distinct dimensions of image assessment and are therefore largely independent. As illustrated in Fig.~\ref{fig:relationship}, an image that receives a high score in IQA may perform poorly in IAA. Likewise, images that achieve strong aesthetic scores may exhibit weaker performance in structural or textural assessment.

\section{Further Discussion on UniPercept}
\label{sec:exploration_unipercept}

\subsection{Training Dataset}
\label{sec:training_dataset}

\noindent\textbf{For Domain-Adaptive Pre-Training.}
As shown in Tab.~\ref{tab:dataset_pretrain}, our Domain-Adaptive Pre-Training is conducted on approximately 800K samples spanning the IAA, IQA, and ISTA domains. For the IAA and IQA domains, the data include both text and rating types: the text data consist of high-quality image--text pairs curated from the source datasets using an MLLM (GPT-4o~\cite{gpt4}), while the rating data are directly taken from ArtiMuse-10K~\cite{artimuse} and KonIQ-10K~\cite{koniq}. For the ISTA domain, due to the lack of high-quality annotations, we first extract images that meet the domain requirements from the raw datasets and then, following the definition in Sec.~\ref{sec:definition_ista}, employ GPT-4o to generate structured annotations, which are further converted into textual descriptions. Using all the above data, we perform Domain-Adaptive Pre-Training to equip UniPercept with an initial capacity for perceptual-level image understanding.

\noindent\textbf{For Task-Aligned RL.}
As shown in Tab.~\ref{tab:dataset_rl}, we summarize the data composition used in the Task-Aligned RL stage. For the three domains (IAA, IQA, and ISTA), we combine data from both the VR and VQA tasks. For the VR task, we use the training sets from ArtiMuse-10K, KonIQ-10K, and ISTA-10K. For the VQA task, we construct training data following the same pipeline described in Fig.~4 of the main paper, and denote the resulting training set as \textit{UniPercept Data-VQA (train)}. By jointly leveraging all six types of training sources, the Task-Aligned RL stage enables UniPercept to effectively handle both \textbf{two} task formats and all \textbf{three} perceptual domains.

\begin{table}[htbp]
    \centering
    \caption{\textbf{Data Overview for Domain-Adaptive Pre-Training.}}
    \label{tab:dataset_pretrain}
    \vspace{-6pt}
    \resizebox{0.99\linewidth}{!}{%
    \begin{tabular}{c | c  c  c}
        \toprule[1.5pt]
        \textbf{Domain} & \textbf{Type} & \textbf{Size} & \textbf{Source} \\
        \midrule

        \multirow{6}{*}{IAA} 
            & \multirow{5}{*}{Text} & \multirow{5}{*}{$\sim$360K}
            & APDDv2~\cite{apddv2} \\
            & & & Impressions~\cite{impressions} \\
            & & & AVA~\cite{ava} \\
            & & & TAD66K~\cite{tad66k} \\
            & & & FLICKR-AES~\cite{flickr} \\
        \cmidrule(lr){2-4}
            & Rating & $\sim$9K 
            & ArtiMuse-10K~\cite{artimuse} \\
        \midrule

        \multirow{8}{*}{IQA} 
            & \multirow{7}{*}{Text} & \multirow{7}{*}{$\sim$380K}
            & Q-Ground-100K~\cite{q-ground} \\
            & & & DQ-495K~\cite{depictqa_v1} \\
            & & & DataDepictQA~\cite{depictqa_v1,depictqa_v2} \\
            & & & SPAQ~\cite{spaq} \\
            & & & KADID~\cite{kadid10k} \\
            & & & PIPAL~\cite{pipal} \\
        \cmidrule(lr){2-4}
            & Rating & $\sim$7K
            & KonIQ-10K~\cite{koniq} \\
        \midrule

        \multirow{7}{*}{ISTA} 
            & \multirow{6}{*}{Text} & \multirow{6}{*}{$\sim$40K}
            & DTD~\cite{dtd} \\
            & & & FMD~\cite{fmd} \\
            & & & Big and Small Objects~\cite{big_and_small_objects} \\
            & & & Scene Size x Clutter Database~\cite{Scene_Size_Clutter_Database} \\
            & & & Reachspaces~\cite{Reachspaces}\\
            & & & Flickr2K~\cite{flickr2k}, LSDIR~\cite{lsdir} \\
        \cmidrule(lr){2-4}
            & Structure & $\sim$40K
            & \textit{Same as Text} \\
        \bottomrule[1.5pt]
    \end{tabular}
    }
\end{table}

\begin{table}[htbp]
    \centering
    \caption{\textbf{Data Overview for Task-Aligned RL.}}
    \label{tab:dataset_rl}
    \vspace{-6pt}
    \resizebox{0.99\linewidth}{!}{%
    \begin{tabular}{c  c   c  c}
        \toprule[1.5pt]
        \textbf{Domain} & \textbf{Type} & \textbf{Size} & \textbf{Source} \\
        \midrule

        \multirow{2}{*}{IAA} 
            & VR  & $\sim$9K  
            & ArtiMuse-10K~\cite{artimuse} \\

            & VQA & $\sim$10K
            & UniPercept Data-VQA (train) \\
        \midrule

        \multirow{2}{*}{IQA} 
            & VR & $\sim$7K
            & KonIQ-10K~\cite{koniq} \\

            & VQA & $\sim$10K
            & UniPercept Data-VQA (train) \\
        \midrule

        \multirow{2}{*}{ISTA} 
            & VR & $\sim$10K
            & ISTA-10K \\

            & VQA & $\sim$10K
            & UniPercept Data-VQA (train) \\
        \bottomrule[1.5pt]
    \end{tabular}
    }
\end{table}

\subsection{Ablation Studies}
\label{sec:ablation}

We further investigate the training of UniPercept and evaluate its performance on both the VR and VQA dimensions of UniPercept-Bench. The results are reported in Tab.\ref{tab:ablation_vr} and Tab.\ref{tab:ablation_vqa}.


\subsubsection{Training Strategy}

\noindent\textbf{Domain-Adaptive Pre-Training.} 
We empirically verify the importance of Domain-Adaptive Pre-Training. As shown in Tab.\ref{tab:ablation_vr} and Tab.\ref{tab:ablation_vqa}, removing Domain-Adaptive Pre-Training leads to a substantial performance drop on both the VR and VQA tasks. This indicates that a vanilla MLLM initialization exhibits limited perceptual-level understanding, and that training on sufficiently large, domain-relevant data is necessary to equip the model with fundamental perceptual capabilities.

\noindent\textbf{Reward Design.} 
For the VR task, we also consider a threshold-based reward, following the formulation used in Q-Insight~\cite{Q-insight}, which determines correctness solely based on the numerical deviation between the prediction and the ground truth. Let $p_i$ and $g_i$ denote the predicted and ground-truth scores. A prediction is considered correct if its absolute error falls within a predefined tolerance $\epsilon$, and incorrect otherwise. This produces a binary reward signal that avoids extreme reward magnitudes and encourages predictions to stay within an acceptable range:
\begin{equation}
r^{(i)}_{\mathrm{thr}} =
\begin{cases}
1, & \text{if } |p_i - g_i| < \epsilon, \\
0, & \text{otherwise}.
\end{cases}
\end{equation}
In our experiments, we set the threshold to match the one used by the Adaptive Gaussian Soft Reward in UniPercept. As shown in Tab.\ref{tab:ablation_vr} and Tab.\ref{tab:ablation_vqa}, the Adaptive Gaussian Soft Reward consistently outperforms the threshold-based formulation and even leads to improvements on the VQA task, where rating rewards are not directly applied. These results highlight the advantages of the Adaptive Gaussian Soft Reward in more accurately capturing the deviation between predicted and ground-truth scores. They also indicate that the VQA and VR tasks share underlying correlations, such that advances in one task can benefit the other.

\subsubsection{Training Data}

\noindent\textbf{Multi-Task vs. Single-Task.} We further investigate the relationship between the VR and VQA tasks. During both Domain-Adaptive Pre-Training and Task-Aligned RL, we separate the VR and VQA training data while keeping all other text data unchanged, and conduct two settings: VQA-only and VR-only. As shown in Tab.\ref{tab:ablation_vr} and Tab.\ref{tab:ablation_vqa}, the VQA-only model performs poorly on the VR task, and the VR-only model similarly performs poorly on the VQA task. Moreover, both settings underperform UniPercept (VR \& VQA) even on their respective target tasks. These results demonstrate that jointly training on both VR and VQA tasks provides substantial mutual benefits and leads to stronger perceptual understanding.

\noindent\textbf{Multi-Domain vs Single-Domain.} We also study the relationships among the three perceptual domains: IAA, IQA, and ISTA. During both Domain-Adaptive Pre-Training and Task-Aligned RL, we separate the training data of the three domains while keeping all other data unchanged, and conduct three single-domain settings: IAA-only, IQA-only, and ISTA-only. As shown in Tab.\ref{tab:ablation_vr} and Tab.\ref{tab:ablation_vqa}, models trained on a single domain perform well on their corresponding domain but fall short in overall performance. In contrast, UniPercept (trained with a mixture of all three domains) achieves substantially stronger overall results and even surpasses the single-domain models on certain tasks. These findings indicate that, although IAA, IQA, and ISTA focus on different aspects of perceptual assessment, jointly training on all three domains enhances the model’s holistic perceptual understanding.

\begin{table}[htbp]
\centering
\small
\caption{\textbf{Ablation studies on UniPercept-Bench-VR.}
The best results are highlighted with \legendlabel{100}{dark blue} cells, and the second-best results with \legendlabel{40}{light blue} cells. Metrics: SRCC/PLCC.}
\vspace{-6pt}
\resizebox{0.49\textwidth}{!}{
\begin{threeparttable}[t]
\renewcommand{\arraystretch}{0.9}

\begin{tabular}{l|cc|cc|c}
\toprule[1.5pt]
\multirow{2}{*}{\textbf{Models}} &
\multicolumn{2}{c|}{\textbf{IAA}} &
\multicolumn{2}{c|}{\textbf{IQA}} &
\multicolumn{1}{c}{\textbf{ISTA}} \\
\cmidrule(lr){2-3} \cmidrule(lr){4-5} \cmidrule(lr){6-6}
& ArtiMuse-10K~\cite{artimuse} & \textbf{Avg.} &
 KonIQ-10K~\cite{koniq} & \textbf{Avg.} &
 ISTA-10K \\
\midrule

\multicolumn{4}{l}{\textit{\textbf{Ablation on Training Strategy}}}  \\
w/ Threshold Reward & 
0.604/0.556 & \bluecellnp{100}{0.617/0.596} &
0.882/0.888 & \bluecellnp{70}{0.801/0.790} &
0.303/0.334 \\

w/o Adaptive Pre-Training &
0.546/0.510 & 0.481/0.421 &
0.851/0.817 & 0.733/0.700 &
0.755/0.732 \\

\midrule
\multicolumn{4}{l}{\textit{\textbf{Ablation on Training Tasks}}} \\
VQA-Only &
0.591/0.582 & 0.585/0.598 &
0.816/0.847 & 0.769/0.774 &
0.206/0.206 \\

VR-Only &
\bluecellnp{70}{0.629/0.596} & 0.558/0.509 &
0.907/0.828 & 0.794/0.749 &
0.767/0.767 \\

\midrule
\multicolumn{4}{l}{\textit{\textbf{Ablation on Training Domains}}} \\
IAA-Only &
0.621/0.608 & 0.508/0.464 &
0.641/0.644 & 0.706/0.680 &
0.197/0.197 \\

IQA-Only &
0.369/0.352 & 0.468/0.435 &
\bluecellnp{70}{0.901/0.839} & 0.786/0.726 &
0.341/0.337 \\

ISTA-Only &
0.351/0.319 & 0.275/0.288 &
0.595/0.575 & 0.611/0.570 &
\bluecellnp{100}{0.771/0.782} \\

\midrule
\textbf{UniPercept (Ours)} &
\textbf{\bluecellnp{100}{0.746/0.738}} &
\textbf{\bluecellnp{70}{0.590/0.586}} &
\textbf{\bluecellnp{100}{0.940/0.949}} &
\textbf{\bluecellnp{100}{0.824/0.827}} &
\textbf{\bluecellnp{70}{0.778/0.767}} \\
\bottomrule[1.5pt]

\end{tabular}
\end{threeparttable}}
\label{tab:ablation_vr}
\end{table}

\begin{table}[htbp]
\centering
\small
\caption{\textbf{Ablation studies on UniPercept-Bench-VQA.}}
\vspace{-6pt}
\resizebox{0.49\textwidth}{!}{
\begin{threeparttable}[t]
\renewcommand{\arraystretch}{0.9}

\begin{tabular}{l|cccc}
\toprule[1.5pt]
\textbf{Experiments} & \textbf{IAA} & \textbf{IQA} & \textbf{ISTA} & \textbf{Avg.} \\
\midrule

\multicolumn{5}{l}{\textit{\textbf{Ablation on Training Strategy}}} \\
w/ Threshold-based Reward          & 72.32\% & \bluecellnp{70}{76.29\%} & 81.65\% & \bluecellnp{70}{76.75\%} \\
w/o Adaptive Pre-Training    & 69.16\% & 75.09\% & 80.00\% & 74.75\% \\

\midrule
\multicolumn{5}{l}{\textit{\textbf{Ablation on Training Tasks}}} \\
VQA-Only             & 71.92\% & \bluecellnp{70}{76.29\%} & 81.44\% & 76.55\% \\
VR-Only              & 68.57\% & 68.38\% & 75.15\% & 70.70\% \\

\midrule
\multicolumn{5}{l}{\textit{\textbf{Ablation on Training Domains}}} \\
IAA-Only           & \bluecellnp{70}{73.69\%} & 69.67\% & 75.57\% & 72.98\% \\
IQA-Only           & 64.73\% & 76.01\% & 77.53\% & 72.76\% \\
ISTA-Only          & 69.56\% & 69.58\% & \bluecellnp{70}{82.27\%} & 73.80\% \\
\midrule
\textbf{UniPercept (Ours)} &
\bluecellnp{100}{\textbf{76.55\%}} &
\bluecellnp{100}{\textbf{81.07\%}} &
\bluecellnp{100}{\textbf{84.23\%}} &
\bluecellnp{100}{\textbf{80.62\%}} \\

\bottomrule[1.5pt]
\end{tabular}

\end{threeparttable}}
\label{tab:ablation_vqa}
\vspace{-15pt}
\end{table}

\section{More Examples of UniPercept-Bench}
\label{sec:example_benchmark}
\begin{figure*}[htbp]
    \centering
    \includegraphics[width=0.99\textwidth]{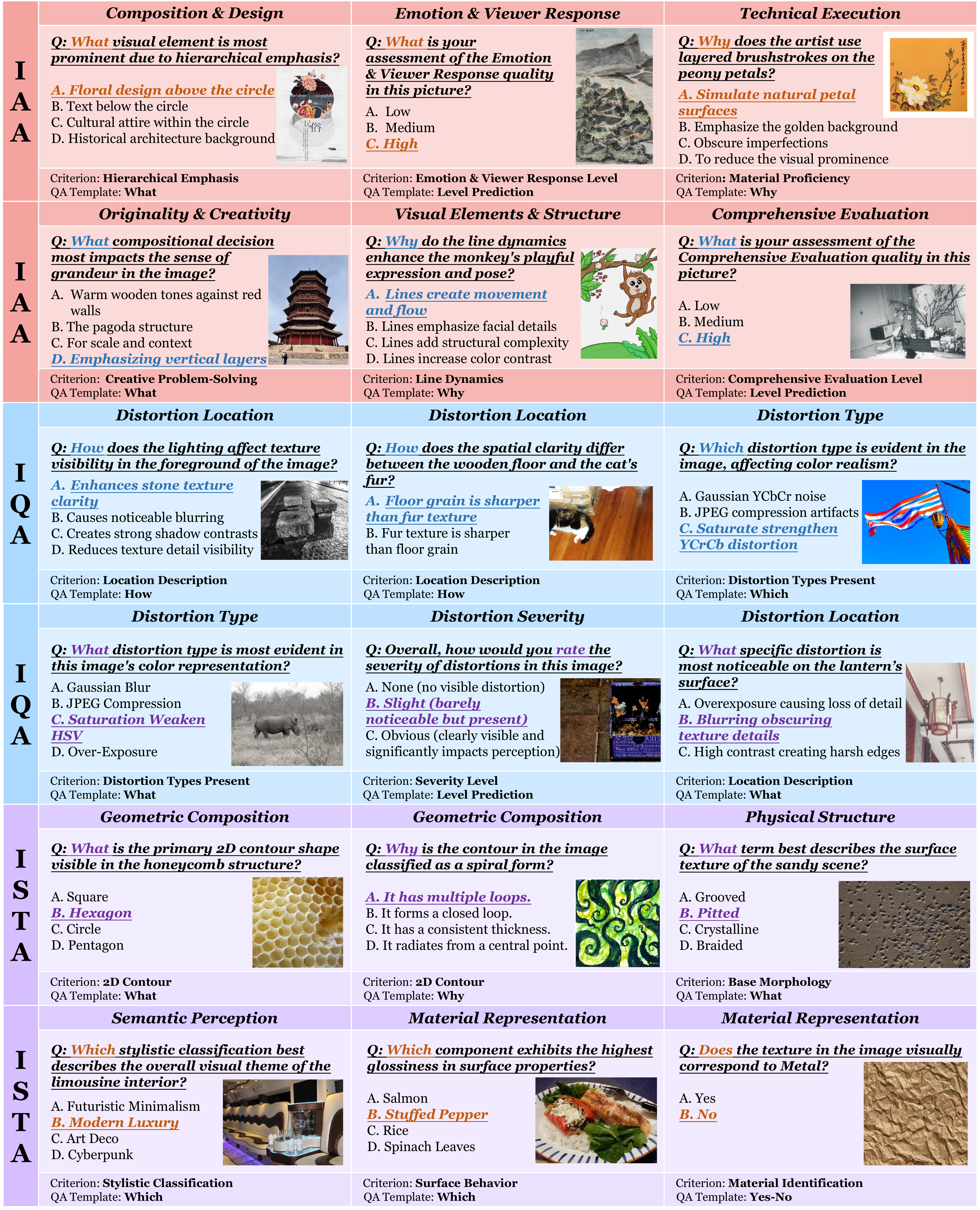}
    \vspace{-10pt}
    \caption{\textbf{More Examples in UniPercept-Bench.}}
    \label{fig:ex_bench_appendix}
\end{figure*}

We provide additional examples from UniPercept-Bench in Fig.~\ref{fig:ex_bench_appendix}.

\section{UniPercept-Constructed Image Profiles}
\label{sec:unifided_unipercept}
\begin{figure*}[htbp]
    \centering
    \includegraphics[width=0.9\textwidth]{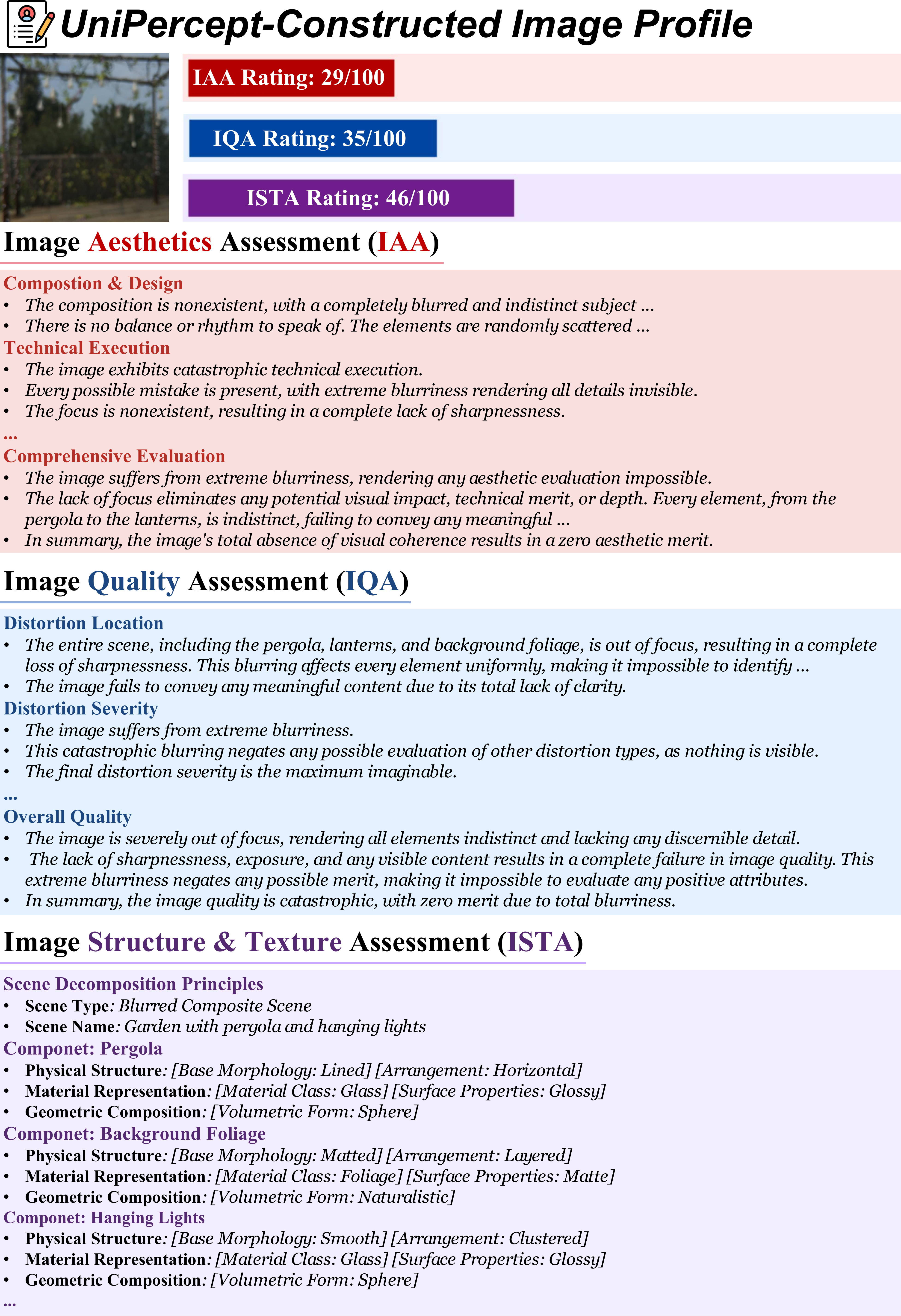}
    \vspace{-10pt}
    \caption{\textbf{Example of \textbf{UniPercept-Constructed Image Profiles}.}}
    \label{fig:ex1}
\end{figure*}

\begin{figure*}[htbp]
    \centering
    \includegraphics[width=0.9\textwidth]{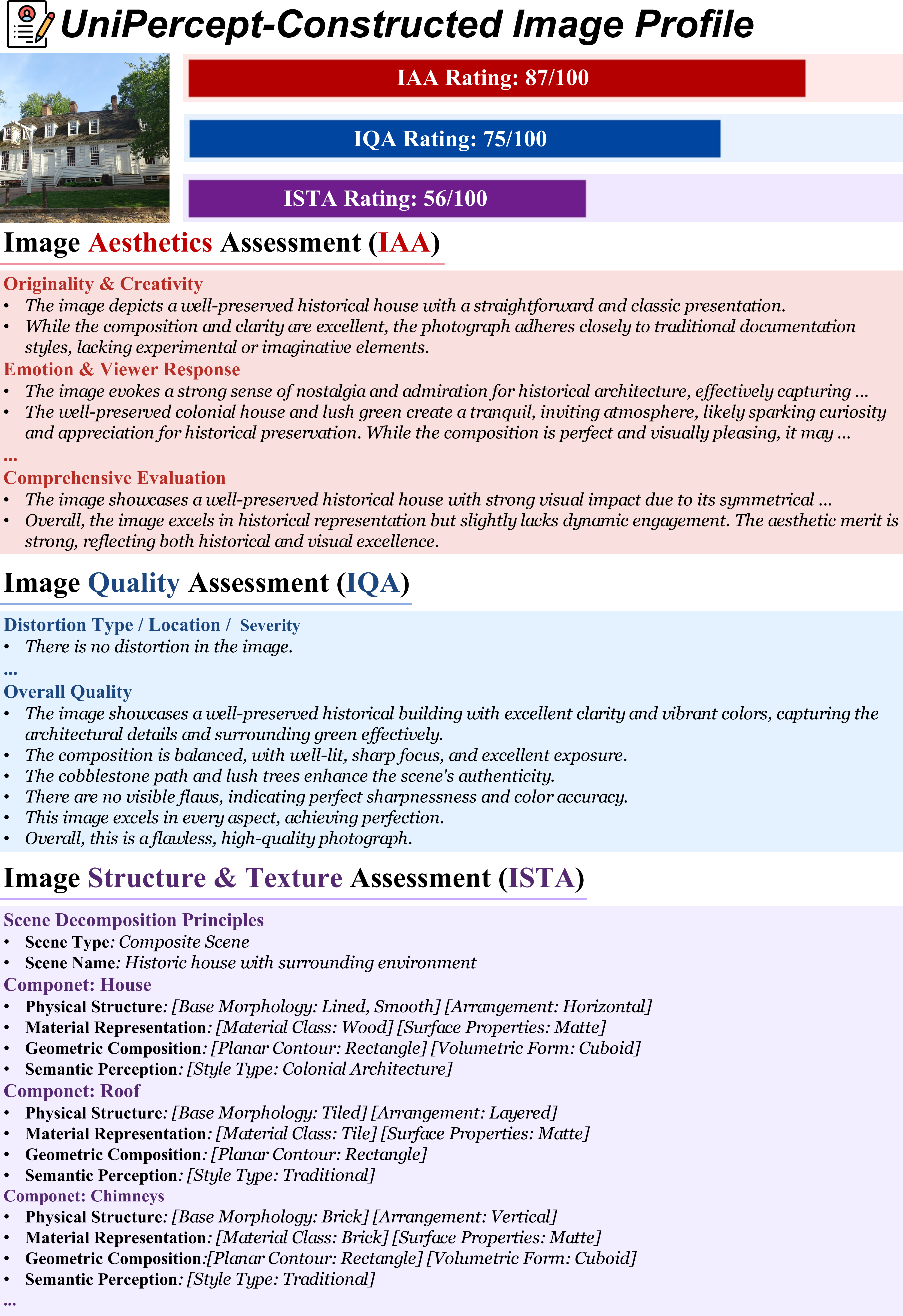}
    \vspace{-10pt}
    \caption{\textbf{Example of \textbf{UniPercept-Constructed Image Profiles}.}}
    \label{fig:ex2}
\end{figure*}

\begin{figure*}[htbp]
    \centering
    \includegraphics[width=0.9\textwidth]{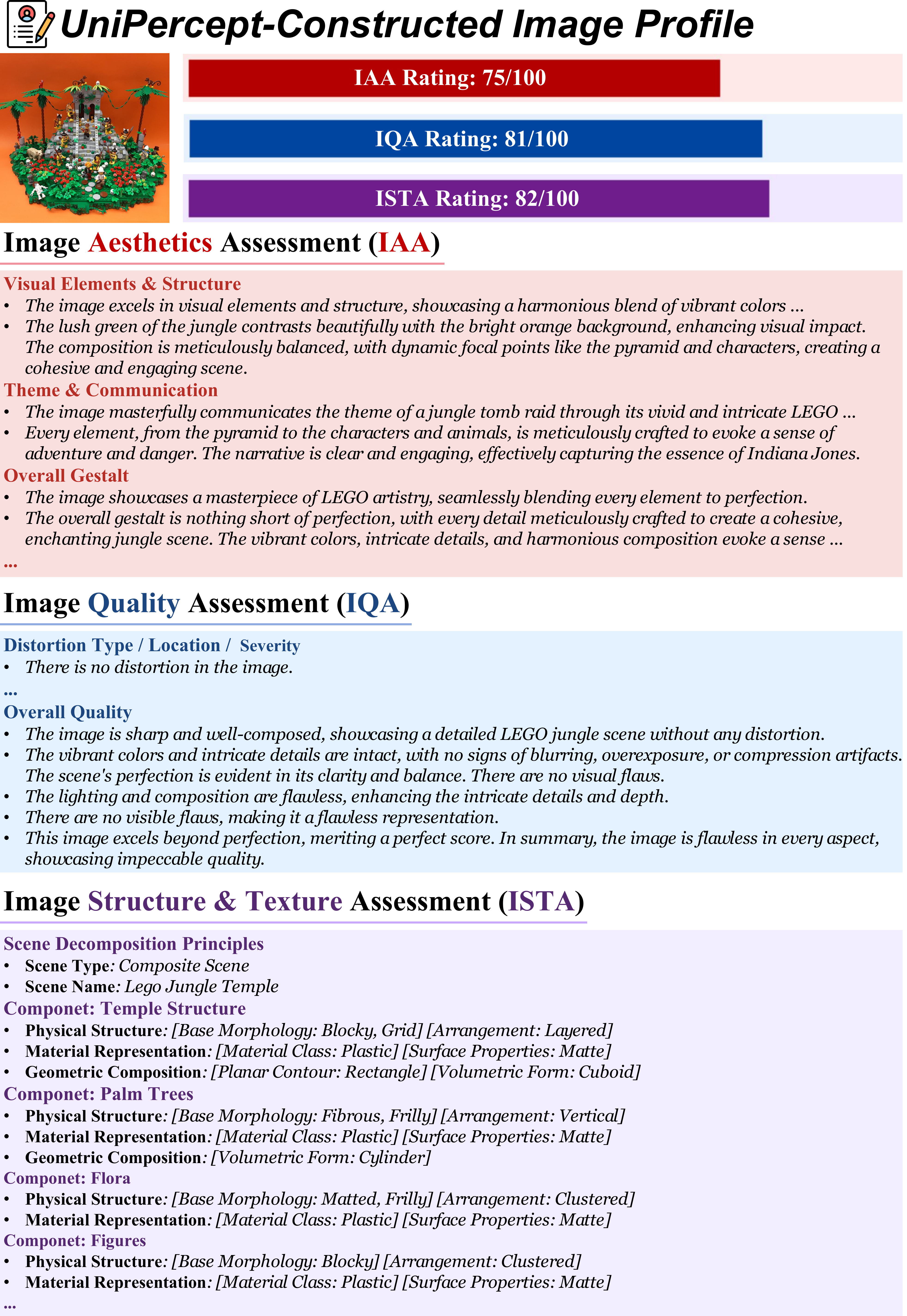}
    \vspace{-10pt}
    \caption{\textbf{Example of \textbf{UniPercept-Constructed Image Profiles}.}}
    \label{fig:ex3}
\end{figure*}

UniPercept is capable of performing comprehensive perceptual-level analysis of images, providing accurate visual-rating evaluations across the IAA, IQA, and ISTA dimensions, together with fine-grained, multi-dimensional analytical outputs. This enables UniPercept to generate a detailed \textit{profile} for each image. We present examples of \textbf{UniPercept-Constructed Image Profiles} in Fig.~\ref{fig:ex1}, Fig.~\ref{fig:ex2}, and Fig.~\ref{fig:ex3}.


\end{document}